\documentclass[journal,10pt]{IEEEtran}
\usepackage[utf8]{inputenc} 
\usepackage{float} 
\usepackage{cite}
\ifCLASSINFOpdf
  \usepackage[pdftex]{graphicx}
  \DeclareGraphicsExtensions{.pdf,.jpeg,.png}
\else
  \usepackage[dvips]{graphicx}
  \DeclareGraphicsExtensions{.eps}
\fi
\usepackage[cmex10]{amsmath}
\usepackage{algorithmic}
\usepackage{array}

\ifCLASSOPTIONcompsoc
 \usepackage[caption=false,font=normalsize,labelfont=sf,textfont=sf]{subfig}
\else
 \usepackage[caption=false,font=footnotesize]{subfig}
\fi
\usepackage{fixltx2e}
\usepackage{stfloats}
\usepackage{url}
\usepackage{amssymb}
\usepackage{graphicx} 
\usepackage{gensymb} 
\usepackage{multirow}

\usepackage{color}

\newcommand{\vect}[1]{\boldsymbol{\mathrm{#1}}}
\newcommand{\vectt}[1]{\boldsymbol{\mathrm{#1}}_t}
\newcommand{\vectall}[1]{\vect{#1}_{1:T}}

\newcommand{\rv}[1]{\textcolor{black}{#1}}

\begin{document}
\title{\textit{GeoTrackNet}---A Maritime Anomaly Detector using Probabilistic Neural Network Representation of \\ AIS Tracks and \textit{A Contrario} Detection}

\author{\IEEEauthorblockN{Duong Nguyen, ~\IEEEmembership{Member,~IEEE,},
Rodolphe Vadaine,
Guillaume Hajduch, \\
Ren\'e Garello, ~\IEEEmembership{Fellow,~IEEE,} and
Ronan Fablet, ~\IEEEmembership{Senior Member,~IEEE}}


\thanks{Duong Nguyen, Ren\'e Garello and Ronan Fablet are with IMT Atlantique, Lab-STICC, 29238 Brest, France (email: \{van.nguyen1, rene.garello, ronan.fablet\}@imt-atlantique.fr)}
\thanks{Rodolphe Vadaine and Guillaume Hajduch are with CLS: Collect Localisation Satellites, 29280 Brest, France (email: \{rvadaine, ghajduch\}@groupcls.com)}

\thanks{This paper is an extension of the MultitaskAIS presented in \cite{nguyen_multi-task_2018}. While \cite{nguyen_multi-task_2018} presents the ability of handling noisy and irregularly sampled data, as well as the computational benefits of this architecture for multiple tasks in maritime surveillance, this paper focuses on detailing the most important task: anomaly detection.}

\thanks{This work was supported by public funds (Minist\`ere de l'Education Nationale, de l'Enseignement Sup\'erieur et de la Recherche, FEDER, R\'egion Bretagne, Conseil G\'en\'eral du Finist\`ere, Brest M\'etropole), by Institut Mines T\'el\'ecom, received in the framework of the VIGISAT program managed by ``Groupement Bretagne T\'el\'ed\'etection'' (BreTel), and by Microsoft (AI EU Ocean awards). It benefited from HPC and GPU resources from Azure (Microsoft EU Ocean awards) and from GENCI-IDRIS (Grant 2020-101030).
The authors acknowledge the support of DGA (Direction G\'en\'erale de l'Armement) and ANR (French Agence Nationale de la Recherche) under reference ANR-16-ASTR-0026 (SESAME initiative),the ANR AI Chair OceaniX, the Labex Cominlabs, the Brittany Council and the GIS BRETEL (CPER/FEDER framework).}

}

\maketitle


\begin{abstract}

Representing maritime traffic patterns and detecting anomalies from them are key to vessel monitoring and maritime situational awareness. We propose a novel approach---referred to as \textit{GeoTrackNet}---for maritime anomaly detection from AIS data streams. Our model exploits state-of-the-art neural network schemes
to learn a probabilistic representation of AIS tracks and \textit{a contrario} detection to detect abnormal events. The neural network provides a new means to capture complex and heterogeneous patterns in vessels' behaviours, while the \textit{a contrario} detector takes into account the fact that the learnt distribution may be location-dependent. Experiments on a real AIS dataset comprising more than 4.2 million AIS messages demonstrate the relevance of the proposed method compared with state-of-the-art schemes. 

\end{abstract}

\section{Introduction}
\label{sec:introduction}

Nowadays, about 90\% of the world trade is carried by maritime traffic, and it is growing consistently \cite{wan_four_2016}. Maritime surveillance and Maritime Situational Awareness (MSA) are vital demands. In this context, anomaly detection is one of the most important tasks, because anomalies may involve accidents (loss of navigation, damages in engine, etc.) or illegal activities (smuggling, illegal transhipment, etc.). Initially designed for collision avoidance, the Automatic Identification System (AIS) has quickly become the main source of information for maritime surveillance, thanks to its information richness. Roughly speaking, AIS messages contain the identity (the MMSI number), the GPS coordinates (latitude, longitude), the current speed (Speed Over Ground--SOG) and course (Course Over Ground--COG), as well as other information about the vessel and the voyage. A series of AIS messages gives the trajectory of the vessel. The potential of AIS is enormous, however, it is not fully utilised. AIS data are awash in noise, besides that, the massive amount of data quickly overwhelms human processing capacity. This emphasises the need for a system that can automatically analyse and arise an alarm whenever there is an abnormal event. However, since AIS was originally created for collision avoidance only, no metadata (quality, reliability, uncertainty, etc.) are available, making the detection of anomalies from AIS a very difficult task. Morever, AIS data in particular, and trajectory data in general, have some specific characteristics that other types of data do not: geographical features, temporal correlations, geographical-temporal features. For these reasons, anomaly detection methods used in other domains such as network traffic analysis or cybersecurity \cite{nanduri_anomaly_2016}, \cite{radford_network_2018} do not apply. We may also emphasise there are no representative groundtruth datasets for this task, hence, supervised learning strategies for anomaly detection as in \cite{song_anomalous_2018, ma_detecting_2018, bouritsas_automated_2019} do not apply either.

Here, we present \textit{GeoTrackNet}---a new approach for maritime trajectory-based anomaly detection\footnote{The detection presented here is trajectory-based, i.e. we focus on the behaviours of vessels. Point-based methods, where the detection is focused on AIS signal, are out of scope of this paper.} using a probabilistic RNN-based (Recurrent Neural Network) representation of AIS tracks and \textit{a contrario} detection. This paper is an extended version of our previous work in \cite{nguyen_multi-task_2018}. The first step in \textit{GeoTrackNet} is to build a normalcy model that represents the characteristics of AIS tracks. At sea, either being enforced by law or for optimisation issues (e.g. optimal fuel consumption, safety purposes, optimal patterns for fishing, etc.), vessels follow some specific patterns, and we expect to learn these patterns from data \cite{mazzarella_discovering_2014, bomberger_associative_2006, pallotta_vessel_2013, arguedas_maritime_2018, dobrkovic_maritime_2018, nguyen_multi-task_2018}. In this work, we exploit variational sequential latent models, specifically the Variational Recurrent Neural Network (VRNN) \cite{chung_recurrent_2015} to create a probabilistic representation of vessels' movement patterns. RNNs have been famous for their ability to capture long-term correlations in time series (here AIS tracks), VRNNs are an extension of RNNs where stochastic factors are added to improve the networks' capacity of modelling data variations and uncertainties. This architecture is one of the state-of-the-art methods for text, speech and music analysis and generation \cite{chung_recurrent_2015, fraccaro_sequential_2016, maddison_filtering_2017}. Besides the quality of AIS signals, which may depend on the metocean conditions as well as interferences in dense traffic areas, vessel trajectory data may also reflect sea surface and wind conditions. These different sources of variations beyond the behavioural patterns of the vessels make anomaly detection in AIS data streams a particularly challenging task.
In this context, VRNNs emerge as a promising candidate for AIS series modelling.  In the proposed scheme, given the learnt representation of the movement patterns of vessels, a ``geospatial \textit{a contrario}" detector evaluates how likely an AIS track segment is to state the detection of abnormal patterns. This detector exploits a geospatial prior depending on the location-dependent complexity of the patterns observed in the considered dataset. This prior also accounts for the strong geographical variabilities of vessels' occurrences and movement patterns. 

Our contributions are as follows:
\begin{itemize}
    \item We propose a new representation of AIS messages for deep neural networks. \rv{This representation aims to highlight the specific route-related characteristic of trajectory data.}
    \item We propose a new method to build a normalcy model for AIS trajectories. This method relies on VRNNs, which can capture variations and uncertainties in AIS tracks to create a probabilistic representation of vessels' trajectories.
    \item We highlight the fact that vessels' behaviours are geospatially-dependent, hence the model representing AIS trajectories shall also be geospatially-dependent. We propose a new anomaly detection method based on this argument. 
    \item We demonstrate the relevance of the proposed scheme with respect to state-of-the-art approaches on a real dataset comprising more than 4.2 million AIS messages.
\end{itemize}

The paper is organised as follows. In Section \ref{sec:related_work}, we give an overview of related work, and analyse the drawbacks of those models. The details of the proposed approach are presented in Section \ref{sec:proposedApproach}. Section \ref{sec:experiments} demonstrates the relevance of \textit{GeoTrackNet} by experiments on real-life data. Conclusions, remaining challenges and future lines of work are discussed in Section \ref{sec:conclusions}.
 
\section{Related work}
\label{sec:related_work}

Recently, there has been a large number of publications related to maritime anomaly detection using AIS. Among them, we can cite \cite{rhodes_maritime_2005, bomberger_associative_2006, laxhammar_anomaly_2008, ristic_statistical_2008, pallotta_vessel_2013, mascaro_anomaly_2014, dafflisio_maritime_2018, kawaguchi_anomaly_2018, forti_anomaly_2019, varlamis_network_2019} and references in \cite{tu_exploiting_2017, riveiro_maritime_2018}. Those methods can be categorized into two groups: rule-based anomaly detection and learning-based anomaly detection. 

The former group defines the abnormal behaviours explicitly and uses a set of rules to state the detection. A large list of such rules can be found in \cite{kazemi_open_2013}. The advantage of this approach is its interpretability. However, it is difficult to define an exhaustive list of abnormal behaviours, and some terminologies such as fast/slow are relative and are hard to implement in operational systems, which may lower their usefulness.

The latter group uses historical data to learn the implicit detection rules. Since no representative groundtruth data are available for maritime anomaly detection, learning-based anomaly detection schemes cannot apply supervised methods like in \cite{song_anomalous_2018, ma_detecting_2018, bouritsas_automated_2019}. Unsupervised learning methods are then preferred \cite{rhodes_maritime_2005, bomberger_associative_2006, pallotta_vessel_2013, arguedas_maritime_2018, forti_anomaly_2019, varlamis_network_2019, zhao_maritime_2019}. Learning frameworks provide a means to overcome the limitations associated with the definition of an exhaustive list of normal/abnormal behaviours. Given the lack of labelled data for the anomalous class, unsupervised schemes naturally arise as the relevant learning strategies.
Due to its flexibility and its ability to apply on a large scale, this second category of approaches has become the dominant approach in maritime anomaly detection \cite{laxhammar_anomaly_2008, pallotta_vessel_2013, varlamis_network_2019, forti_anomaly_2019}.

Learning-based methods consist of two main stages: i) learning a  normalcy model, ii) detecting deviations from the normalcy. In the first stage, density-based spatial clustering techniques, especially DBSCAN \cite{ester_density-based_1996}, have been very popular \cite{pallotta_vessel_2013, coscia_multiple_2018, dafflisio_detecting_2018, varlamis_network_2019}.
Typically, DBSCAN is applied to cluster the critical points of AIS tracks into so-called Waypoints (WPs): ENs---where vessels enter the Region of Interest (ROI), EXs---where vessels exit the ROI, and POs---where vessels stop. From these WPs, these approaches build a graph whose nodes are the WPs and edges are the maritime routes. Using a probabilistic setting, e.g., Kernel Density Estimation (KDE) \cite{pallotta_vessel_2013}, Gaussian Mixture Models (GMM) \cite{laxhammar_anomaly_2008}, multiple Ornstein-Uhlenbeck (OU) processes \cite{forti_anomaly_2019}, a normalcy model is fitted for each edge. 
The next stage evaluates how likely a new AIS track is in order to state the detection. This is typically achieved by applying a threshold on the distance to the centroid feature vector representing the route \cite{varlamis_network_2019} or on the probability of the AIS track given the normalcy model \cite{pallotta_vessel_2013}, or through an adaptive hybrid Bernoulli filter \cite{forti_anomaly_2019}. 

In all of the above mentioned  methods, the extraction of WPs is critical. However, the considered clustering techniques, such as DBSCAN, may be sensitive to hyper-parameters. Different settings may lead to very different outcomes. Moreover, it is not always possible to link a track to an edge of the normalcy graph, i.e. we can not assign the beginning point and the end point of a track to any WP. This is a common problem of any method based on clustering. Another important limitation of the above mentioned approaches is that they apply to cargo and tanker vessels, and may not apply to other vessel types, for instance, fishing vessels whose AIS patterns do not involve route-like patterns. As AIS metadata may not be reliable, dealing with all vessel types in operational systems would require additional preprocessing steps to filter out vessels' types.

\rv{Although over the last decade, deep learning has achieved very impressive results in many complicated tasks and has become the state-of-the-art approach in many domains \cite{lecun_deep_2015}, AIS-based maritime surveillance is not one of them. Popular network architectures for time series modelling and analysis such as Recurrent Neural Network (RNN), Long-Short Term Memory (LSTM), etc. may hardly model the dynamics of AIS trajectories because the data are noisy and may be effected by external factors (e.g. metocean conditions).} Another issue is that those methods assume that the performance of the learnt model is geospatially-homogeneous. However, in some areas, there are a lot of vessels and their behaviours are similar, the maneuvering patterns in these areas can be learnt easily. By contrast, other areas may involve much less training data and/or highly-complex and multi-modal patterns, which result in poor performance of the learnt normalcy model and of the associated anomaly detection schemes. The application of the same anomaly detection policy (threshold, filter) in these two types of areas does not seem relevant. 

In this paper, we present a new method, referred to as \textit{GeoTrackNet} that tackles those problems by exploiting advances in probabilistic neural network representations for  time series analysis and an \textit{a contrario} detection framework for maritime anomaly detection from AIS data streams. Our method provides a new means to address key issues of state-of-the-art approaches, both in terms of the extraction and representation of the normalcy and of the detection of deviations from the normalcy for all types of vessels.

\section{Proposed Approach}
\label{sec:proposedApproach}

In this section, we present the details of the proposed approach. 
\textit{GeoTrackNet} relies on the architecture of the Embedding layer we introduced for the MultitaskAIS network presented in \cite{nguyen_multi-task_2018}. We first introduce this architecture, then detail the formulation of the proposed anomaly detection method.

\subsection{Data representation}
\label{sec:fourhot}

\begin{figure}
  \centering
  \includegraphics[width=80mm]{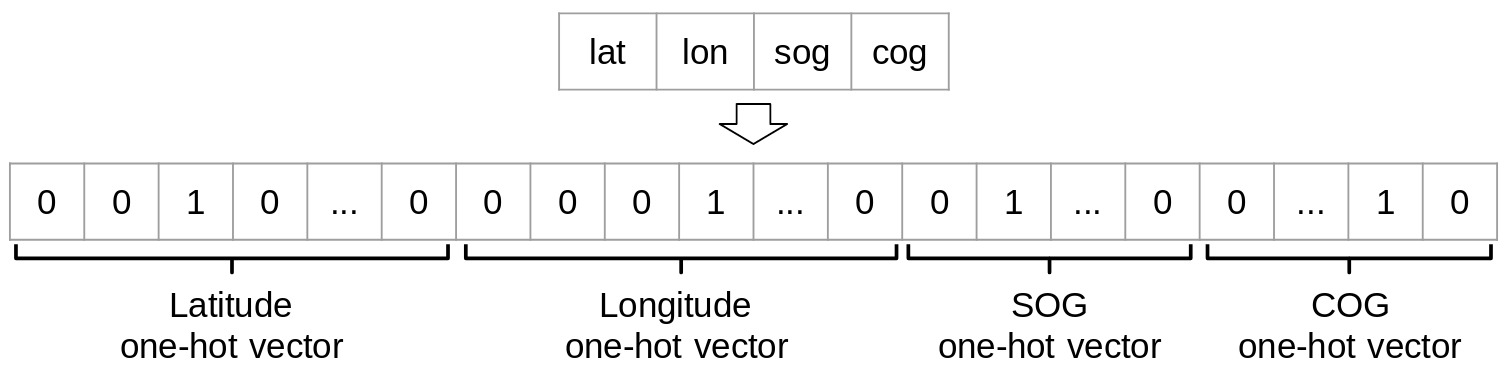}
  \centering
  \caption{``Four-hot'' vector.} \label{figFourHotVector}
\end{figure}

The most common way to represent an AIS message is using a 4-D real-valued vector (two dimensions for the position and the other two for the velocity, e.g. $\left[ lat,lon,SOG,COG \right]^T$) \cite{pallotta_vessel_2013, dafflisio_maritime_2018, forti_anomaly_2019, uney_data_2019}. \rv{We argue that this representation is not suitable for neural-network-based methods, because it is difficult for a neural network to disentangle the underlying geospatial meaning of these numbers. Instead, we represent each AIS point by a ``four-hot vector'' (Fig. \ref{figFourHotVector}). A ``four-hot'' representation is a concatenated vector of the one-hot vectors of the latitude coordinate, longitude coordinate, SOG and COG.}

\begin{figure}
  \centering
  \includegraphics[width=82mm]{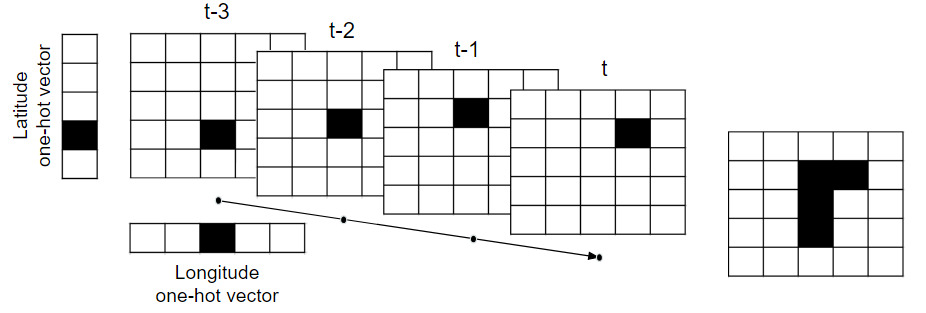}
  \caption{Geometric feature obtained by concatenating the one-hot vectors of the latitude and the longitude coordinates of AIS messages.} \label{figOnehotAccum}
\end{figure}

\begin{figure}
  \centering
  \includegraphics[width=82mm]{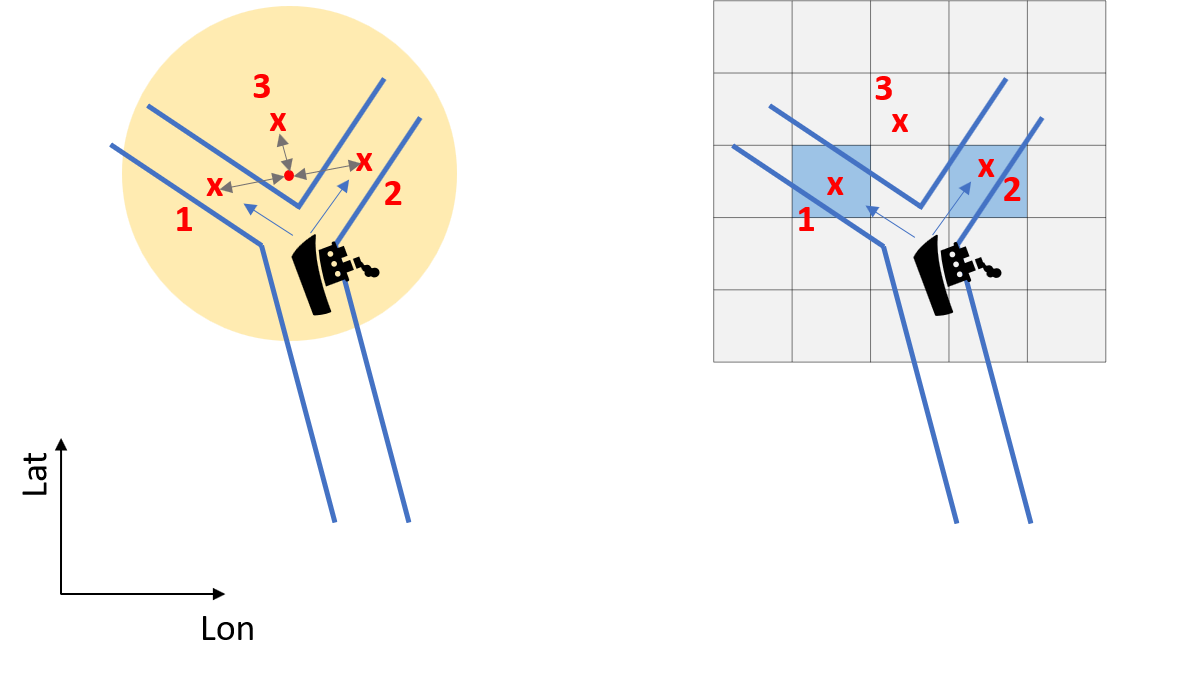}
  \caption{\rv{Continuous real-valued representation (left) vs. ``four-hot" representation (right) of AIS messages in the considered learning-based setting. For the sake of simplicity, SOG and COG are not considered here. Assume that there is a maritime route (depicted by blue lines), and at the junction, half of the vessels in the historical dataset turned left (to position $\vect{x}^1$) and half turned right (to position $\vect{x}^2$), but none of them went straight ahead (to position $\vect{x}^3$). \textbf{\textit{Left}}: If vessel positions are represented by real-valued vectors and the dynamics of vessels are modeled by Gaussian distributions, at the next timestep, the abnormal position $\vect{x}^3$ would yield a better score than the actual normal positions $\vect{x}^1$ and $\vect{x}^2$, because $\vect{x}^3$ is closer to the red dot---the center of the Gaussian distribution (depicted by the yellow circle). \textbf{\textit{Right}}: If vessels positions are represented by ``four-hot" vectors and the dynamics of vessels are modeled by multivariate Bernoulli distributions, at the next timestep, the model would give higher probability values to the two blue ```bins" only, and position $\vect{x}^3$ would be very unlikely compared with positions $\vect{x}^1$ and $\vect{x}^2$}} \label{fig:4hotExplained}.
\end{figure}

In addition to the classically-expected benefits of bucketing representation \cite{bengio_representation_2013}, ``four-hot" vectors help disentangle the geometric features as well as the phase (time-space) patterns of AIS tracks. For example, Fig. \ref{figOnehotAccum} shows how this representation accentuates the geometric feature of an AIS track. Similarly, the phase feature appears when we sum up the one-hot vectors of the latitude, longitude coordinates and the speeds in the resulting 3-D space (see \cite{nguyen_multi-task_2018}). We also expect that during the learning process, the ``four-hot" representation 
enforces route-related characteristics of trajectory data in general, and of AIS data in particular. More precisely, the model shall learn that some vessels should follow some specific routes, and hence detects as abnormal any vessel deviating from the maritime route that it is on. As an illustration, Fig. \ref{fig:4hotExplained} shows how the ``four-hot" representation can help the model detect abnormal movements deviating from maritime routes.

The hyper-parameters are the resolution of each bin in the one-hot vectors. If the resolution is too high, the whole network becomes too bulky and requires a high computational resource to run, and may also lead to overfitting. If the resolution is too low, we may lose critical information. For anomaly detection, we may not need very accurate position and velocity features. For example, a speed of 10 knots or 10.1 knots is not expected to make any difference in the context of anomaly detection. Overall, our experiments suggest that the resolutions of 0.01$\degree$ for longitude and latitude, 1 knot for SOG and 5$\degree$ for COG work well most of the time.

\subsection{Probabilistic Recurrent Neural Network Representation
of AIS Tracks}
\label{sec:vrnn}

In this section we introduce a probabilistic neural network architecture that we use to represent AIS tracks: the Variational Recurrent Neural Network (VRNN) \cite{chung_recurrent_2015}. \rv{VRNNs are an extension of RNNs where stochastic factors are added to improve the modelling capacity of the network. VRNNs are widely used to create generative models for speech \cite{serban_multiresolution_2016}, text \cite{gupta_deep_2017, zaheer_latent_2017}, videos \cite{he_probabilistic_2018}, machine translation \cite{su_variational_2018}, or even physical processes \cite{ajay_augmenting_2018}}. We detail the associated probabilistic formulation and the resulting; however, we present a different derivation which would clarify some terms used in the next sections of this paper. 

For any contiguous AIS track\footnote{A contiguous AIS track is a track whose the time gap between any two successive messages is smaller than a threshold, here 2h.}, we can always apply an interpolation and sampling technique to create a sequence of $T$ variables: $\vectall{x} = \{\vect{x}_t\},_{t=1:T}$, with $\vect{x}_t$ is the ``four-hot" vector representation of AIS messages presented in Section \ref{sec:fourhot}. The objective is to learn a distribution that   maximise the log likelihood $\log p(\vectall{x})$, which can factorise as:
\begin{equation}
    \label{eq:rnn}
    \log p(\vectall{x}) = \log p(\vect{x}_1) \sum_{t=2}^T \log p(\vect{x}_t|\vect{x}_{1:t-1}).
\end{equation}

Recently, Recurrent Neural Networks (RNNs) have emerged as the state-of-the-art approach for time series modelling and analysis \cite{lecun_deep_2015, goodfellow_deep_2016}. RNNs assume that at a given time $t$, the relevant historical information in $\vect{x}_{1:t-1}$ can be encoded in a deterministic hidden state $\vect{h}_t$: $p(\vect{x}_t|\vect{x}_{1:t-1}) = p(\vect{x}_t|\vect{h}_t)$. The dynamics of the series are modeled by a deterministic differentiable function $f$: $\vect{h}_t = f(\vect{x}_{t-1},\vect{h}_{t-1})$. $f$ is usually parameterised by LSTMs \cite{hochreiter_long_1997} or GRUs \cite{chung_gated_2015}. The initial condition $\vect{h}_1$ is commonly set to $\vect{0}$. Eq. \eqref{eq:rnn} becomes: 
\begin{equation}
    \label{eq:p_x_rnn}
    \log p(\vectall{x}) = \sum_{t=1}^T \log p(\vect{x}_t|\vect{h}_t).
\end{equation}{}

The fact that $f$ is deterministic makes RNNs hardly be able to capture all the variations and uncertainties in data. In our context, $f$ can be interpreted as a model of the maneuvering patterns of AIS trajectories. Associated uncertainties may come from AIS data streams themselves as well as their discretisation using ``four-hot" vectors. Variations in AIS data streams may relate to vessel types, weather conditions, AIS message corruption, etc.

To account for such variations and uncertainties, probabilistic RNNs relate to the introduction of latent stochastic variables, denoted as $\vect{z}_t$, which 
follow a prior distribution: 
\begin{equation}
    \vect{z}_t \sim p(\vect{z}_t|\vect{h}_t).
\end{equation}
The dynamics and the emission distribution become:
\begin{equation}
    \vectt{h} = f(\vect{x}_{t-1},\vect{z}_{t-1},\vect{h}_{t-1}),    
\end{equation}
\begin{equation}
    \vectt{x} \sim p(\vectt{x}|\vectt{z},\vectt{h}).
\end{equation}
At each time step $t$, the joint probability of $\vectt{x}$ and $\vectt{z}$ can factorise as:
\begin{equation}
    \label{eq:pt_joint}
    p(\vectt{x},\vectt{z}|\vectt{h}) = p(\vectt{x}|\vectt{z},\vectt{h})p(\vectt{z}|\vectt{h}).
\end{equation}
Hence, $p(\vectt{x}|\vectt{h})$ can be obtained by integrating out $\vectt{z}$ in Eq. \eqref{eq:pt_joint}:
\begin{equation}
    p(\vectt{x}|\vectt{h}) = \mathbb{E}_{p(\vectt{z}|\vectt{x},\vectt{h})} \left[ p(\vectt{x}|\vectt{z},\vectt{h})p(\vectt{z}|\vectt{h}) \right].
\end{equation}
However, this integral is usually intractable. 
Variational approaches propose that instead of maximising $\log p(\vectt{x}|\vectt{h})$, we   maximise a lower bound of this distribution, called the Evidence Lower BOund (ELBO), using an approximation $q(\vectt{z}|\vectt{x},\vectt{h})$ of the true posterior distribution $p(\vectt{z}|\vectt{x},\vectt{h})$ \cite{chung_recurrent_2015, bishop_pattern_2006}:
\begin{multline}
    \mathcal{L}(\vectt{x}|\vectt{h},p,q) = \mathbb{E}_{q(\vectt{z}|\vectt{x},\vectt{h})}\left[\log p(\vectt{x}|\vectt{z},\vectt{h}) \right] \\
    - \mathrm{KL} \left[ q(\vectt{z}|\vectt{x},\vectt{h}) ||  p(\vectt{z}|\vectt{h}) \right]. 
\end{multline}
where $\mathrm{KL} \left[ q(\vectt{z}|\vectt{x},\vectt{h}) ||  p(\vectt{z}|\vectt{z}) \right]$ is the Kullback-Leibler divergence between two distributions $q$ and $p$. 

Overall, given the neural network parameterisation for function $f$, the emission distribution $p(\vectt{x}|\vectt{z},\vectt{h})$ and the approximated posterior distribution $q(\vectt{z}|\vectt{x},\vectt{h})$, the training step comes to   maximise Eq. \eqref{eq:p_x_rnn} where the term $\log p(\vectt{x}|\vectt{h})$ is approximated by $\mathcal{L}(\vectt{x}|\vectt{h},p,q)$. This maximisation is implemented using a stochastic gradient ascent technique. The details of the considered neural network parameterisations for the different building blocks of the model (using LSTMs) are presented in Section \ref{sec:experiments}. 

\subsection{\textit{A contrario} detection}
\label{sec:contrario}

Once distribution $p(\vectall{x})$ is learnt, we can simply apply a ``global thresholding" rule to state the detection, i.e.  AIS tracks whose $\log p(\vectall{x}) < \varepsilon$ are flagged as abnormal, like in our previous work \cite{nguyen_recurrent_2019}. However, as mentioned in Section \ref{sec:related_work}, vessels' behaviours vary significantly, depending on the considered geographical areas. In some areas, AIS tracks may involve multimodal but well-defined patterns, and the learnt model can precisely capture these patterns. As a result, normal AIS tracks shall be associated with high probability values, whereas tracks will low probability values shall relate to unusual and possibly abnormal ones. In other areas, because of the variabilities of vessels' behaviours, limited amount of AIS data and/or a lower capacity of the model to represent AIS tracks, the learnt model may result in low probability values whatever the tracks. In such cases, the use of a global thresholding approach might lead to poorly relevant detection results.

To address these issues, we introduce a new detection method, referred to as ``geospatial \textit{a contrario}" detection. It takes into account the geospatially-heterogeneous performance of the learnt model. We rely on the division of the ROI into a grid. Let us denote by $l_{\vectt{x}}^{C_i}$ the log probability $\log p(\vect{x}_t|\vectt{h})$ of AIS messages in a small geographical cell $C_i$ (i.e., $\vect{x}_t \in C_i$) and $p^{C_i}$ the distribution of $l_{\vectt{x}}^{C_i}$:
\begin{equation}
    l_{\vect{x}_t}^{C_i} \sim p^{C_i}.
\end{equation}
An AIS message in cell $C_i$ is considered as abnormal if its log probability is smaller than the lowest $\frac{1}{p}$-quantile of $p^{C_i}$. 
\begin{equation}
    \vect{x}_t \textit{ is abnormal} \Leftrightarrow  p^{C_i}(\mathrm{L} < l_{\vectt{x}}^{C_i}) < p. 
\end{equation}
That means, if we randomly sample $l_{\vectt{x}}^{C_i}$ from $p^{C_i}$ (note that $p^{C_i}$ is the distribution of variable $l_{\vectt{x}}^{C_i}$, and not $\vectt{x}$), the probability that ``$\vectt{x}$ is abnormal" is $p$. 

Assuming that the event ``$\vect{x}_t$ is abnormal" of each AIS message $\vect{x}_t$ in an AIS track $\boldsymbol{\mathrm{x}}_{1:T}$ is independent, the probability that ``at least $k$ out of $n$ AIS messages in an AIS segment of length $n$ (denoted $\vect{x}_{t:t+n-1}$) of this track are abnormal" is a tail of a Binomial distribution:
\begin{equation}
    \mathcal{B}(n,k,p) = \sum_{i=k}^{n}{n \choose i}p^{i}(1-p)^{n-i}.
\end{equation}
\textit{A contrario} detection \cite{desolneux_gestalt_2008} detects whether such an AIS segment is abnormal based on the Number of False Alarms ($\mathrm{NFA}$), defined as:
\begin{equation}
    \mathrm{NFA}(n,k,p) = N_s\mathcal{B}(n,k,p),
\end{equation}
where $N_s = \frac{T(T+1)}{2}$ is the number of all possible segments.

For example, if $T=3$, there are 6 possible segments: 3 segments of length 1, 2 segments of length 2 and 1 segment of length 3. 

If the $\mathrm{NFA}$ of a track segment is  smaller than a predefined threshold $\varepsilon$, this segment will be considered as abnormal. An AIS track is abnormal if at least one of its segments is abnormal:
\begin{equation}
    \label{eq:thresholding}
    \vectall{x} \textit{ is abnormal}. \\
    \Leftrightarrow  \exists (n,k), \mathrm{NFA}(n,k,p) <  \varepsilon . 
\end{equation}
The threshold $\varepsilon$ is the allowed expectation of ``false alarm", that means, if we run the detector on a series of random $l_{\vect{x}_t}^{C_i}$ $1/\varepsilon$ times, there will be 1 segment flagged as abnormal. Interested readers are referred to \cite{desolneux_gestalt_2008} for more details. To implement this {\em a contrario} scheme, we use two approaches to model distribution $p^{C_i}$: i) a simple Gaussian approximation and ii) a Kernel Density Estimation (KDE) \cite{rosenblatt_remarks_1956},\cite{parzen_estimation_1962}.    
\section{Experiments and results}
\label{sec:experiments}

\subsection{Experimental set-up}

\begin{figure}[!t]
  \centering
  	\subfloat[]
    {\includegraphics[width=70mm]{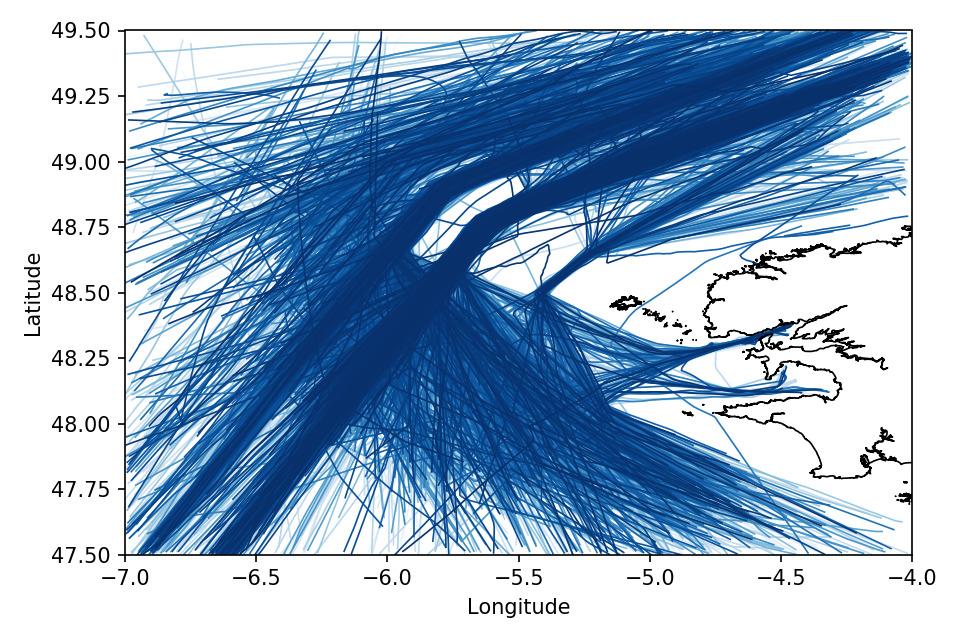}
 	\label{fig:Vs_train}}%
  \hfil
  	\subfloat[]
    {\includegraphics[width=70mm]{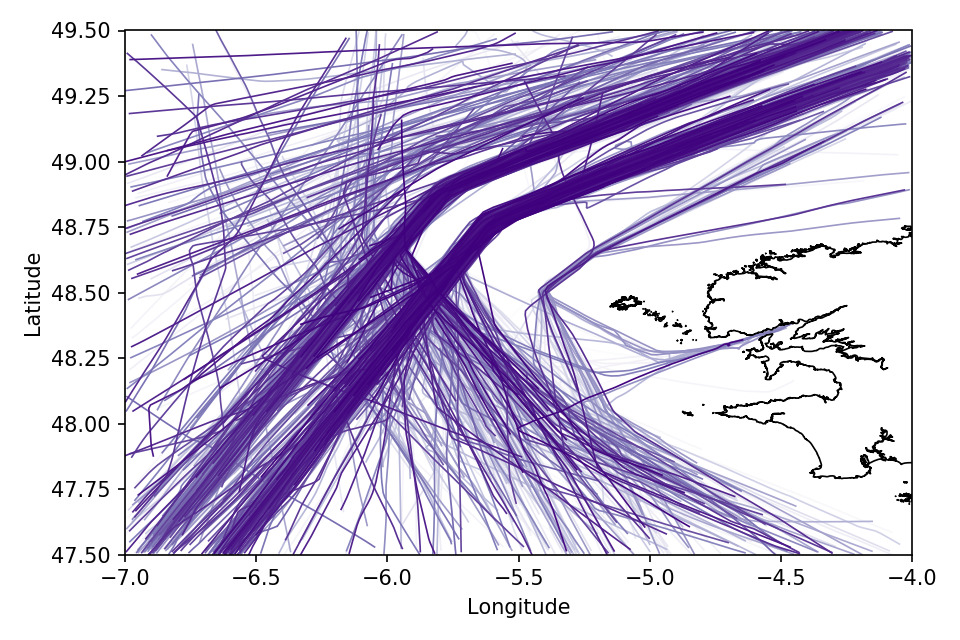}
 	}%
  \hfil
  	\subfloat[]
    {\includegraphics[width=70mm]{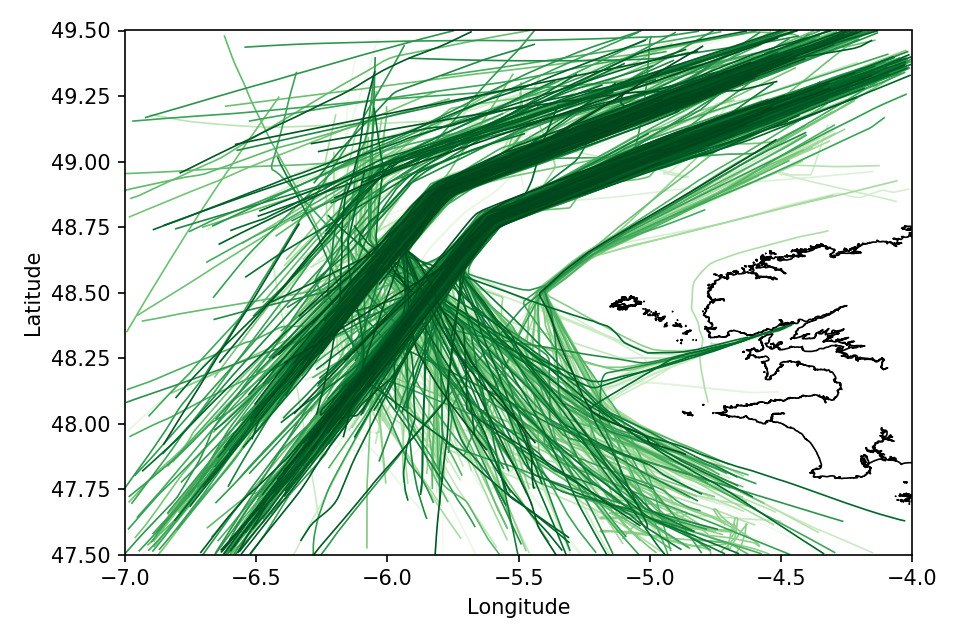}
 	}%
  \centering
  \caption{All AIS tracks in the dataset from January 1 to March 31, 2017. (a) training set; (b) validation set; (c) test set.}
  \label{fig:dataset2017010203}
\end{figure}

\textbf{Datasets}: We tested our model on AIS data received by an AIS station located in Ushant. The ROI was a rectangle from (47.5\degree N, 7.0\degree W) to (49.5\degree N, 4.0\degree W). The data were collected from January to March 2017 and from July to September 2017. In each period, there were more than 4.2 million AIS messages. For each period, we divided the data into three sets: a training set, from the first day to the 10th of the last month of this period (e.g.  from January 1 to March 10); a validation set, from the 11th of the last month to the 20th of the last month (e.g.  from March 11 to March 20) and a test set, from the 21st of the last month to the last day of this period (e.g.  from March 20 to March 31). The basic idea behind this experimental setting is that for operational applications, we use historical data to train the model (i.e. to learn $p(\vect{x}_{1:T})$), then apply the learnt model to current data. The validation sets are used to check for overfitting and for the estimation of distribution $p^{C_i}$. Fig. \ref{fig:dataset2017010203} shows an illustration of the training set, the validation set and the test set of the period from January to March 2017.

\textbf{Preprocessing:} \textit{GeoTracknet} can process AIS streams in real-time. In real-time operational applications, whenever an AIS message arrives, it will be grouped into a track keyed by the MMSI. The detection starts if the track is long enough to be meaningful, here greater or equal to 4 hours. The system incrementally updates the tracks by adding arriving AIS messages and discarding old data. The implementation and the performance of the online detection version of \textit{GeoTracknet} can be found in \cite{nguyen_detection_2020}. Those technical details are out of scope of this paper. Here, for the sake of simplicity, we present the offline version of \textit{GeoTracknet}.

We removed erroneous position or speed messages in the considered AIS data streams. The SOG was truncated to 30 knots. Discontiguous voyages (voyages that have the maximum interval between two successive AIS messages longer than a threshold, here 2 hours) were split into contiguous ones. We re-sampled all voyages to a resolution of 10 minutes (i.e. ,  $\{t+1\} - \{t\} = 10$mins) using a linear interpolation. Very long voyages were split into smaller tracks from 4 to 24 hours each.

\textbf{Neural Network architectures}: for the model reported in this paper, the resolutions of the latitude, longitude, SOG and COG were set to 0.01\degree  (about 1km), 0.01\degree, 1 knot and 5\degree, respectively.  We modeled $f$ by an LSTM with one single hidden layer of size 100 for datasets comprising only cargo and tanker vessels, and of size 120 for datasets comprising all types of vessels. $\vectt{z}$ was real-valued vectors of the same size of the hidden layer of the LSTM. $p(\vectt{z}|\vectt{h})$ and $q(\vectt{z}|\vectt{x},\vectt{h})$ were two Gaussian distributions parameterised by two fully connected networks with one hidden layer of size 100. $p(\vectt{x}|\vectt{h},\vectt{z})$ is a multivariate Bernoulli distribution parameterised by a fully connected network with one hidden layer of size 100. The network was trained using Adam optimiser \cite{kingma_adam:_2015} with a learning rate of 0.0003.

\textbf{\textit{A contrario} detection}: for the \textit{a contrario} detector, we chose $p=0.1$. $\varepsilon$ was initially set at a high value (in order to flag many tracks as abnormal), then was gradually decreased to reduce the number of false positives while keeping all the true detections.  

The code, as well as the data that can replicate the results in this paper are available at: https://github.com/CIA-Oceanix/GeoTrackNet.

\textbf{Baseline}: We used the Traffic Route Extraction and Anomaly
Detection (TREAD) method, presented in \cite{pallotta_vessel_2013, arguedas_maritime_2018} as the baseline. This model supposes that vessels following the same route have similar velocity in each small area. The hyper-parameters were set at the values suggested by \cite{pallotta_vessel_2013} and \cite{varlamis_network_2019} ($minPts = 10$, $eps = 2000$, the radius of each small area is 3 km). We also included state-of-the-art NN models for sequential data, namely  LSTMs \cite{marchi_novel_2015, marchi_non-linear_2015} and VRNNs \cite{nguyen_recurrent_2019, su_robust_2019}.    

\textbf{Evaluation method}: As no reference groundtruth dataset is available, a quantitative benchmarking synthesis in terms of accuracy or false alarm rate is not feasible. We rather analyse the different types of anomalies identified by different models. Besides, a more thorough analysis has been performed for \textit{GeoTrackNet} through an inspection of each detected anomaly by AIS experts.  

\subsection{Experiments and results}

\begin{figure}[!t]
  \centering
  	\subfloat[]
    {\includegraphics[width=65mm]{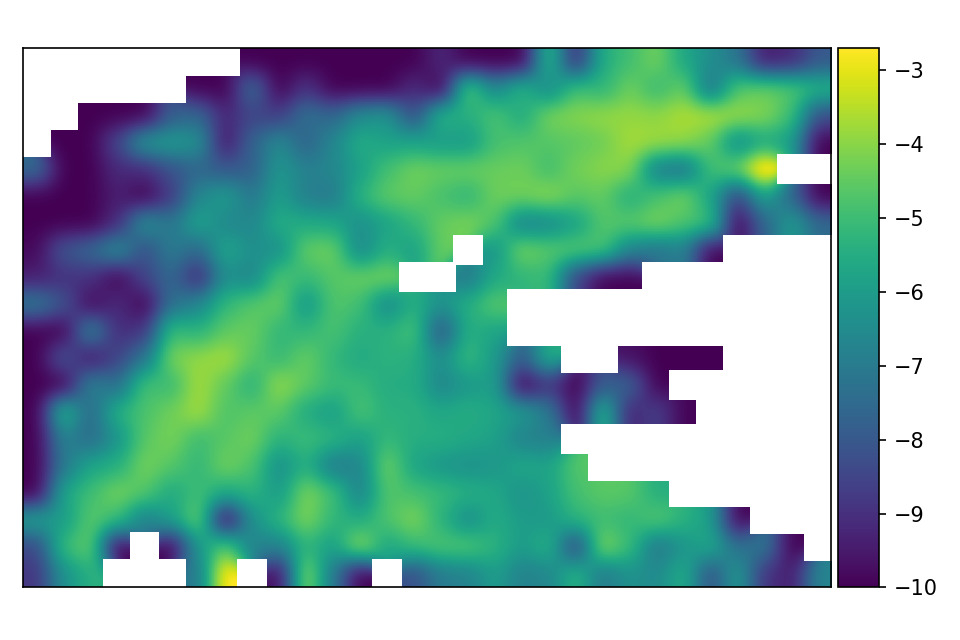}
 	\label{fig:ll_mean}}%
  \hfil
  	\subfloat[]
    {\includegraphics[width=65mm]{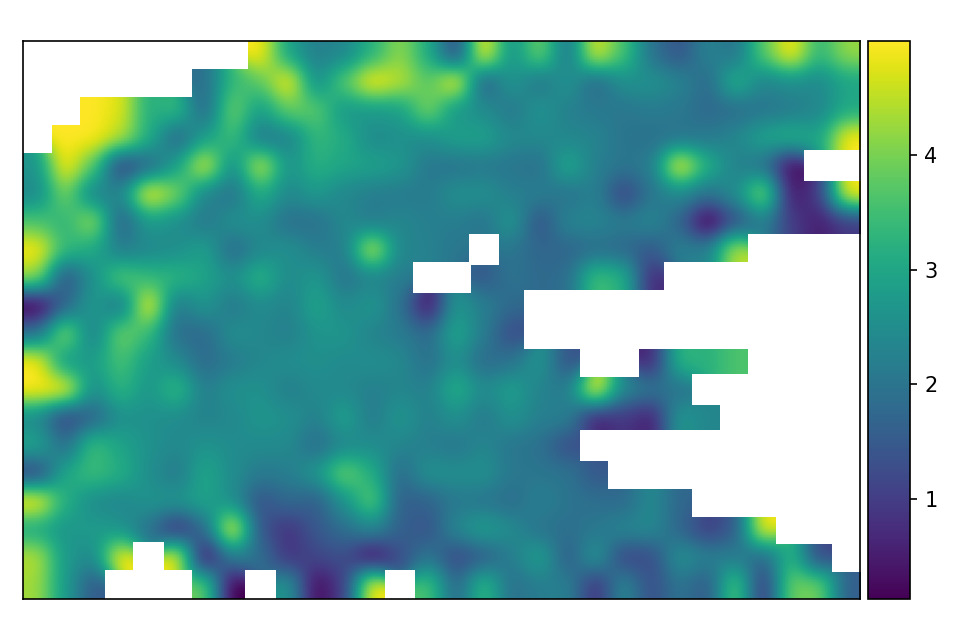}
 	\label{fig:ll_std}}%
  \centering
  \caption{The ``geospatial performance" map displaying the mean (a) and the standard deviation (b) of the Gaussian approximation of distributions $p^{C_i}$ from AIS messages in the validation set from January to March, 2017. On maritime routes, there are many vessels, mainly cargoes and tankers, their movement patterns can be learnt easily, $\log p(\vectt{x}|\vectt{h})$ is usually high and its variation is small. On the other hand, some areas depict few vessels or vessels' behaviours are too complicated for the model to learn, $\log p(\vectt{x}|\vectt{h})$ is usually low and highly variable. Blank regions are regions where we do not apply the detection (e.g., land areas or regions where we do not have enough data).}
  \label{figPerformance}
\end{figure}

\textbf{Basic case study}: For this test, we trained the models on the training set and evaluated the performance on the corresponding test set of each period. The dataset comprises only cargoes and tankers. Fig. \ref{figPerformance} shows the mean and the standard deviation of distributions $p^{C_i}$. As expected, in some regions, there are many vessels and the learnt model fits well the data with a mono-modal or multimodal distribution, such that the values of $\log p(\vectt{x}|\vectt{h})$ are high. There are also regions where  $\log p(\vectt{x}|\vectt{h})$ is  low on average. If an AIS track results in a low log probability in these regions, we do not know whether this track is unusual or the model does not fit well the data. Applying a ``global thresholding" rule  like in \cite{nguyen_recurrent_2019} would lead to a bad outcome, as shown in Fig. \ref{fig:global_thresholding}, where all the detections are in low log likelihood regions. By contrast, the proposed \textit{a contrario} detector compares $\log p(\vectt{x}|\vectt{h})$ of an AIS message $\vectt{x}$ with those in the same area, if it is significantly smaller than the others, then $\vectt{x}$ is regarded as abnormal. The results are shown in Fig. \ref{fig:ct_gauss} and Fig. \ref{fig:ct_kde}. Most of the time, the model using Gaussian distribution approximation and the one using KDE gives similar outcomes. The proposed model can detect both: i) space-wise (geometric and geographic) anomalies, when vessels deviate from maritime routes, perform unusual turns, etc. and ii) phase-wise (kinetic) anomalies, when vessels have abnormal evolution in speed and course (e.g.  unusual slowing down, sudden changes in speed, etc.), as shown in Fig. \ref{fig:each_anomaly}. Among those 25  tracks flagged as abnormal in Fig. \ref{fig:ct_kde}, AIS experts reported only one (the dark yellow track turning north at (49\degree N, 5\degree W)) as a false alarm.
We suspect this detection relates to the low number of AIS tracks in this area in the training set as this area is outside of the coverage zone of the terrestrial AIS station located in Ushant. Additional experiments reported in \cite{nguyen_detection_2020} support this statement as the model trained with a larger training set (comprising both terrestrial AIS and satellite AIS) does not flag this AIS track as abnormal. 

\rv{Regarding LSTM and VRNN models (Fig. \ref{fig:LSTM} and Fig. \ref{fig:VRNN}, respectively), the performance does not appear very relevant. They flag many normal tracks as abnormal. For example, in both figures, the tracks along the 6.0$\degree$W longitude line are usual tracks. In Fig. \ref{fig:VRNN}, the yellow, orange and red tracks departing from Brest (48.4\degree N, 4.5\degree W) are normal tracks (except the red track in Fig. \ref{fig:transhipment}).}

When comparing our approach to TREAD \cite{pallotta_vessel_2013}, we note that some types of anomaly, such as the double U-turn, abnormal turns, or abnormal speeds, are detected by both approaches, as shown in Fig. \ref{fig:TREAD} and Fig. \ref{fig:ct_kde}. However, since TREAD compares the velocity of a vessel with the average of vessels on the same route to state the detection, this method is sensitive to vessels' speed. TREAD considers all vessels that move slower or faster than others as abnormal. This may lead to some unexpected results, when the statistical anomaly is not suspicious, like the one in Fig. \ref{fig:speedcom_a}. This vessel is flagged by TREAD because it moved too fast. However, it might not involve any suspicious activity. On the other hand, \textit{GeoTrackNet} focuses more on sudden changes in speed of vessels, see Fig. \ref{fig:speedcom_d} for an example. This detection is relevant because this vessel may encounter an engine failure.

\newcommand{\sfsizea}{59mm}%
\begin{figure*}
  \centering
    \subfloat[]
    {\includegraphics[width=\sfsizea]{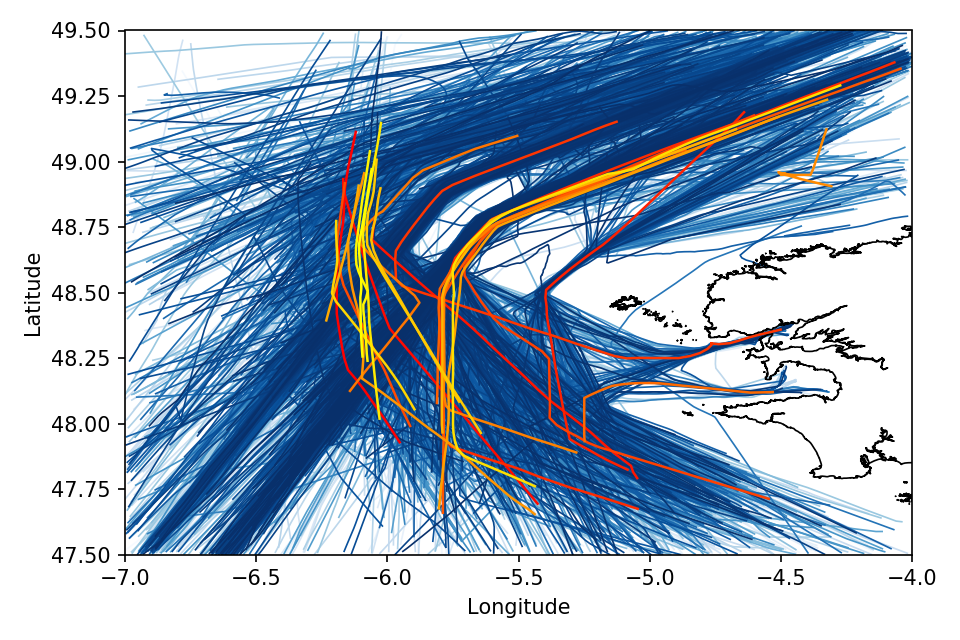}
 	\label{fig:LSTM}}%
    \hfil
    \subfloat[]
    {\includegraphics[width=\sfsizea]{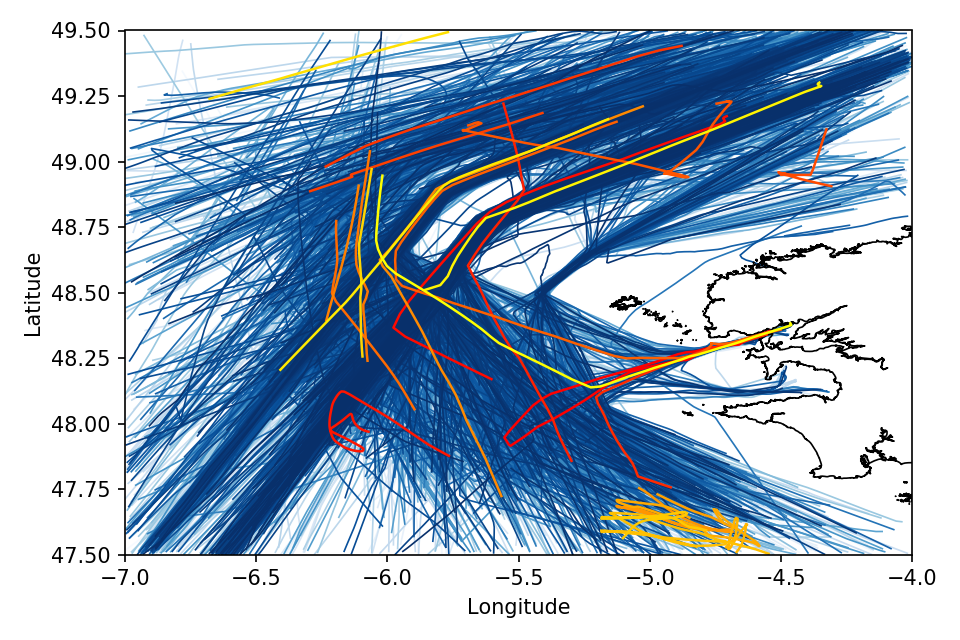}
 	\label{fig:VRNN}}%
    \hfil
  	\subfloat[]
    {\includegraphics[width=\sfsizea]{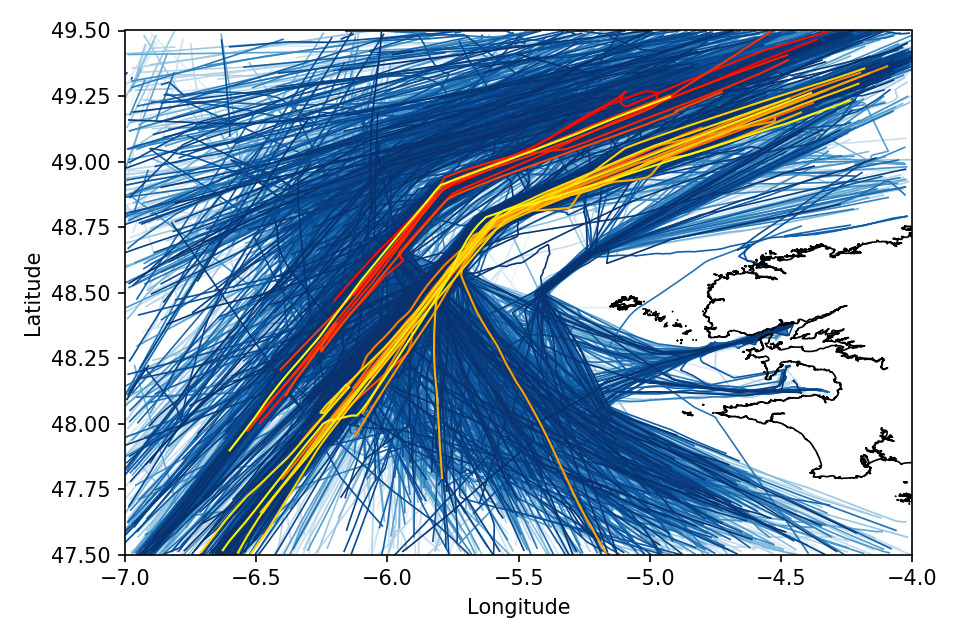}
 	\label{fig:TREAD}}%
    \hfil
  	\subfloat[]
    {\includegraphics[width=\sfsizea]{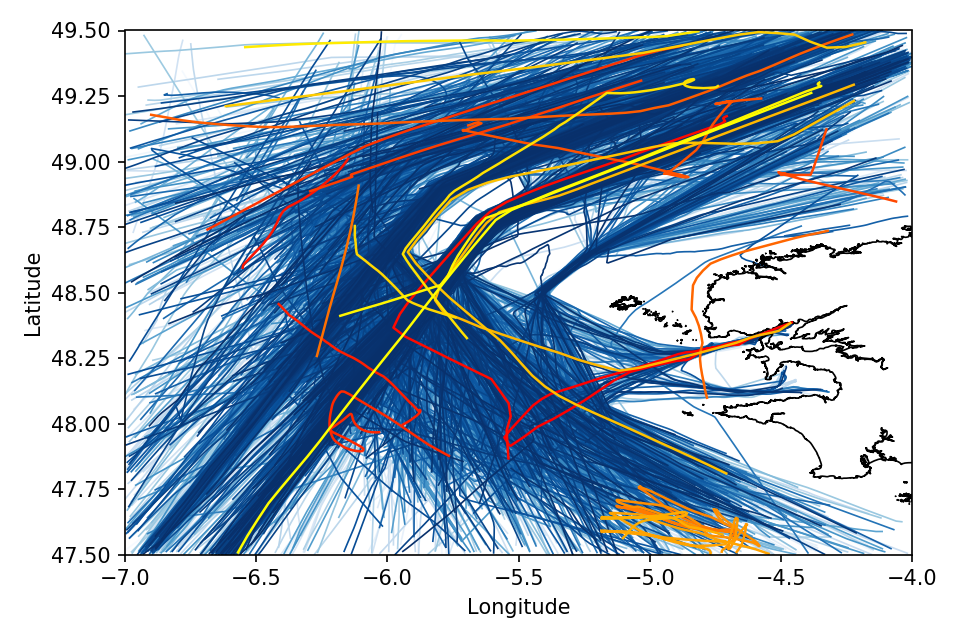}
 	\label{fig:global_thresholding}}%
    \hfil
  	\subfloat[]
 	{\includegraphics[width=\sfsizea]{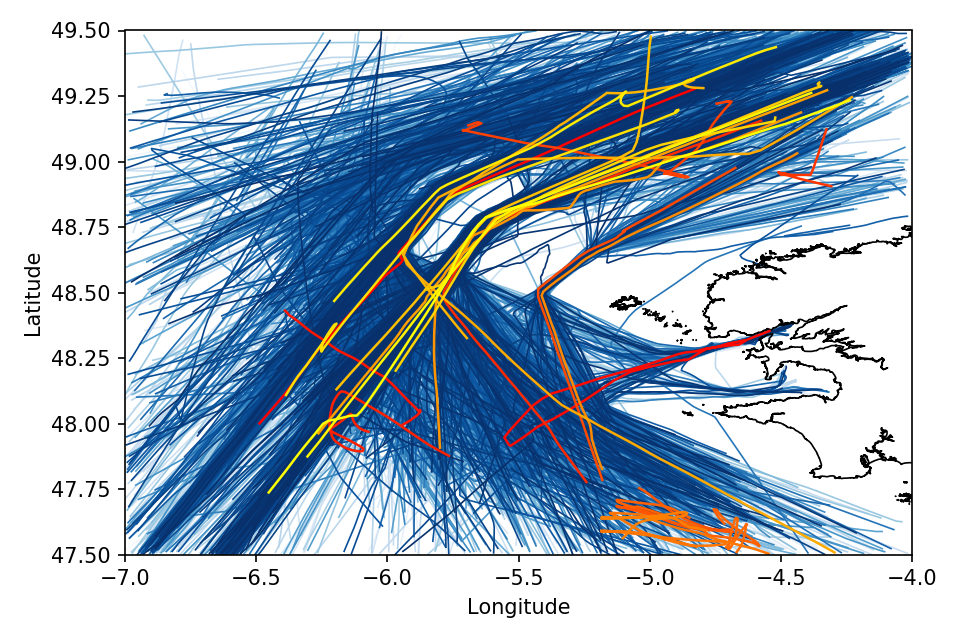}
 	\label{fig:ct_gauss}}%
    \hfil
  	\subfloat[]
    {\includegraphics[width=\sfsizea]{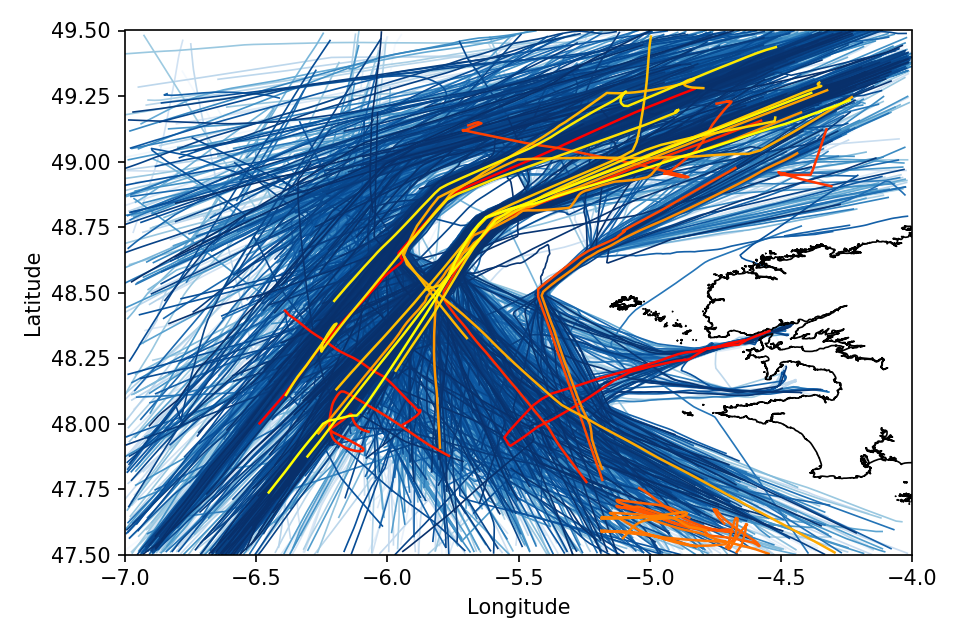}
 	\label{fig:ct_kde}}%
  \centering
  \caption{\rv{Abnormal tracks detected by different models (the dataset comprises only cargo and tanker vessels, from January to March 2017). Blue: tracks in the training set; other colors: abnormal tracks in the test set (the colors of abnormal tracks were chosen randomly). (a) LSTM; (b) VRNN, (c) TREAD (a DBSCAN-based method introduced in \cite{pallotta_vessel_2013}); (d) \textit{GeoTrackNet} without the \textit{a contrario} detector (i.e. using a ``global thresholding" rule); (e) \textit{GeoTrackNet}, approximating each $p^{C_i}$ by a Gaussian distribution; and (f) \textit{GeoTrackNet}, approximating each $p^{C_i}$ by KDE.}}
    \label{fig:ct_2017010203}
\end{figure*}

\newcommand{\sfsize}{59mm}%
\begin{figure*}
  \centering
  	\subfloat[]
    {\includegraphics[width=\sfsize]{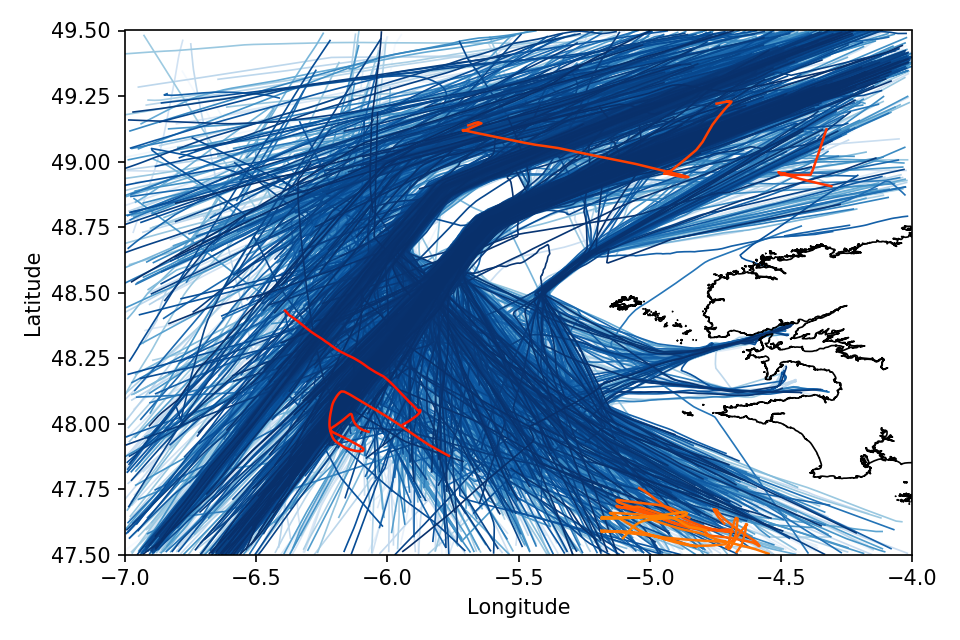}
 	\label{fig:graph}}%
  \hfil
  	\subfloat[]
    {\includegraphics[width=\sfsize]{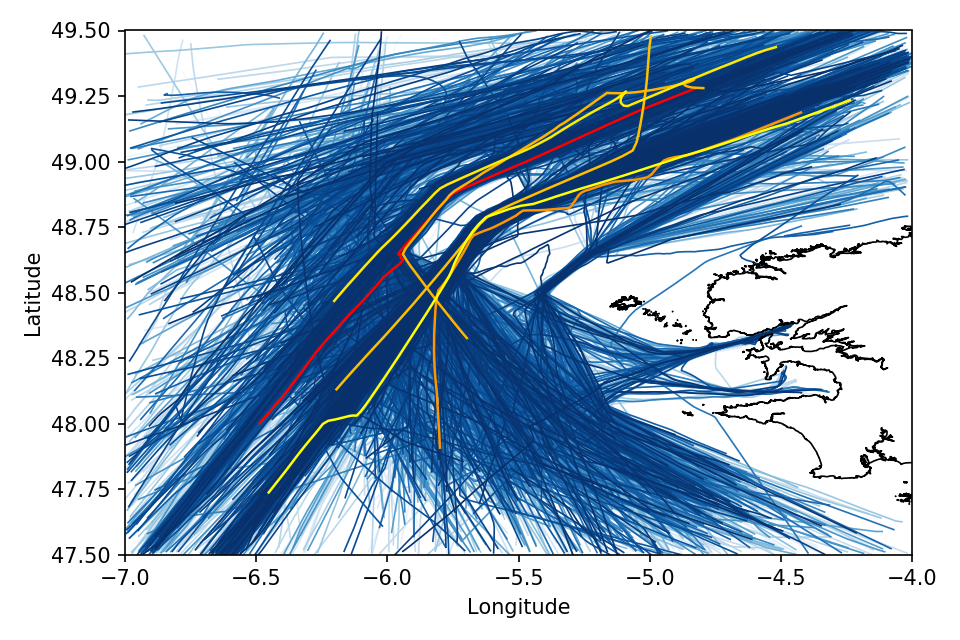}
 	\label{fig:turns}}%
  \hfil
  	\subfloat[]
    {\includegraphics[width=\sfsize]{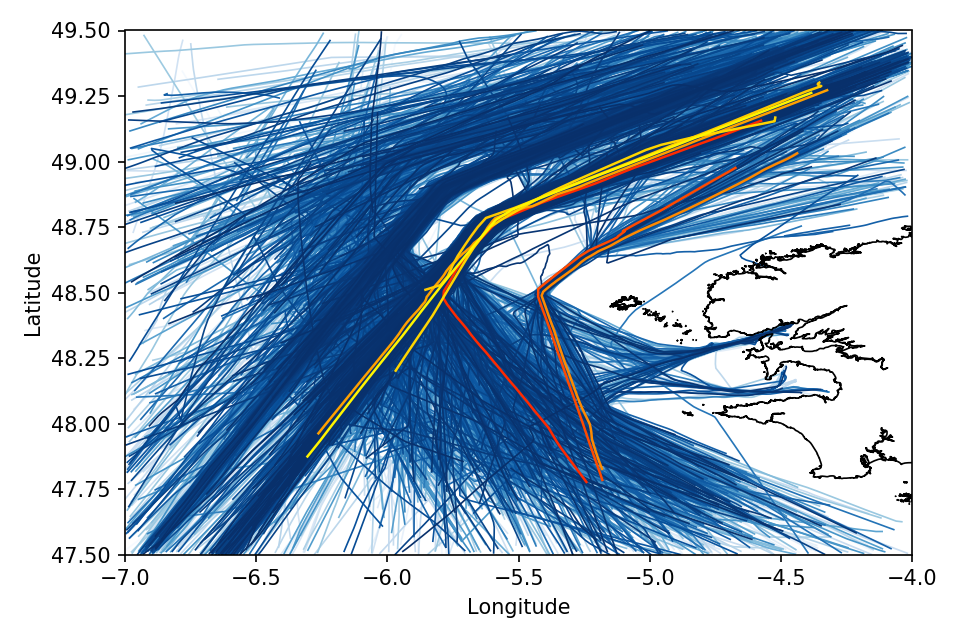}
 	\label{fig:speed}}%
  \hfil
  	\subfloat[]
    {\includegraphics[width=\sfsize]{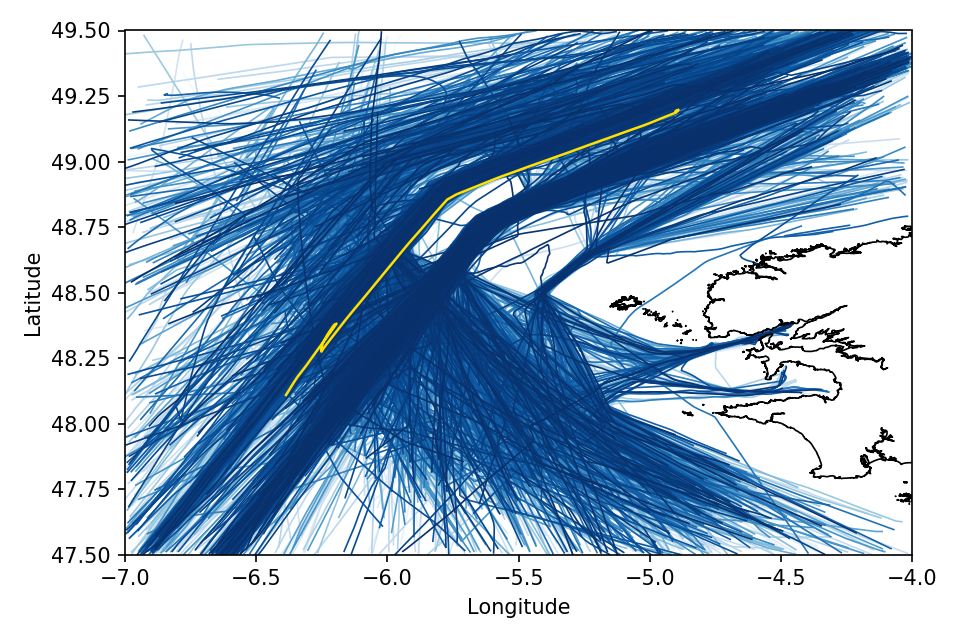}
 	\label{fig:2Uturn}}%
  \hfil
  	\subfloat[]
    {\includegraphics[width=\sfsize]{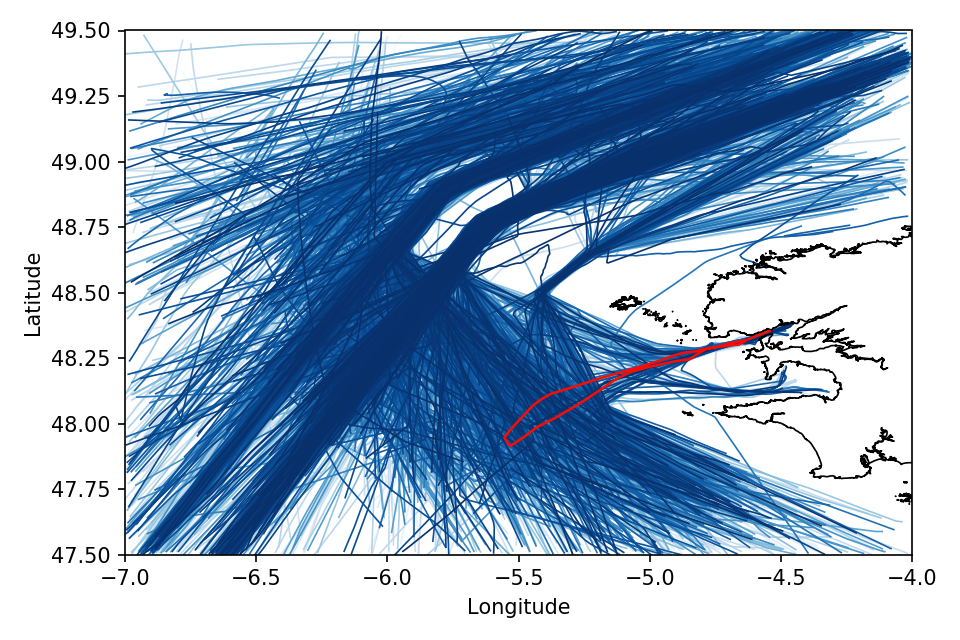}
 	\label{fig:transhipment}}%
  \hfil
  	\subfloat[]
    {\includegraphics[width=\sfsize]{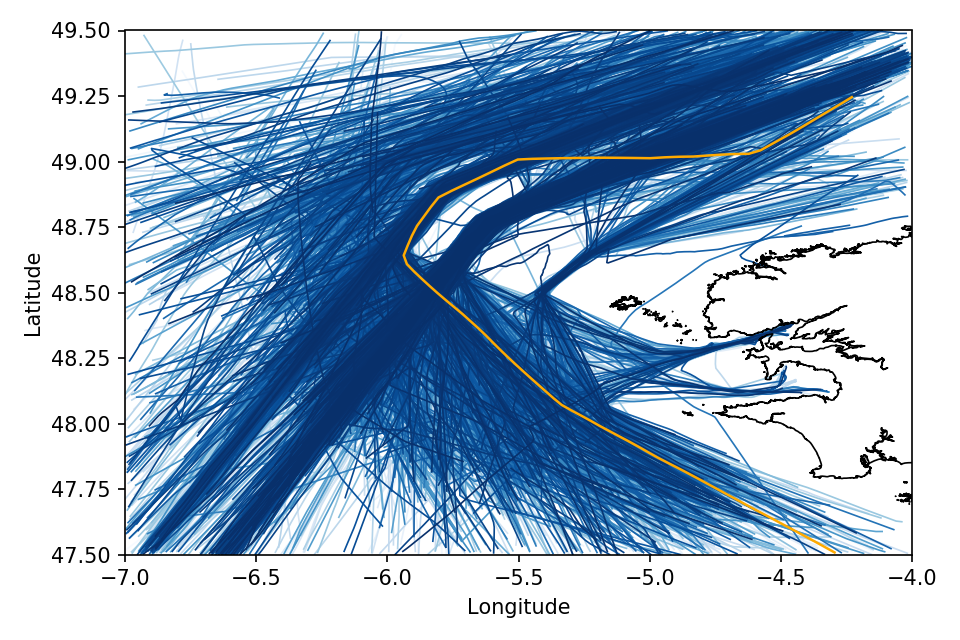}
 	\label{fig:kde}}%
  \centering
  \caption{Examples of anomalies detected by KDE \textit{GeoTrackNet}. (a) Vessels following abnormal routes. DBSCAN-based methods can not apply to these tracks because they can not be assigned to any common maritime route. (b) Geometrically or geographically abnormal tracks (e.g., deviating from maritime routes, unusual turns, etc.). (c) Abnormal speed tracks (e.g. suspiciously slowing down in a maritime route). (d) Double U-turn. (e) A cargo vessel steamed to sea then went back to the departing port. (f) Each segment of this track is normal, however, it is unusual that a vessel follows this path. \textit{GeoTrackNet} can detect this track because it has a memory (the memory of its LTSM).
  \label{fig:each_anomaly}}
\end{figure*}

\newcommand{\sfsizec}{43mm}%
\begin{figure}
  \centering
  	\subfloat[]
    {\includegraphics[width=\sfsizec]{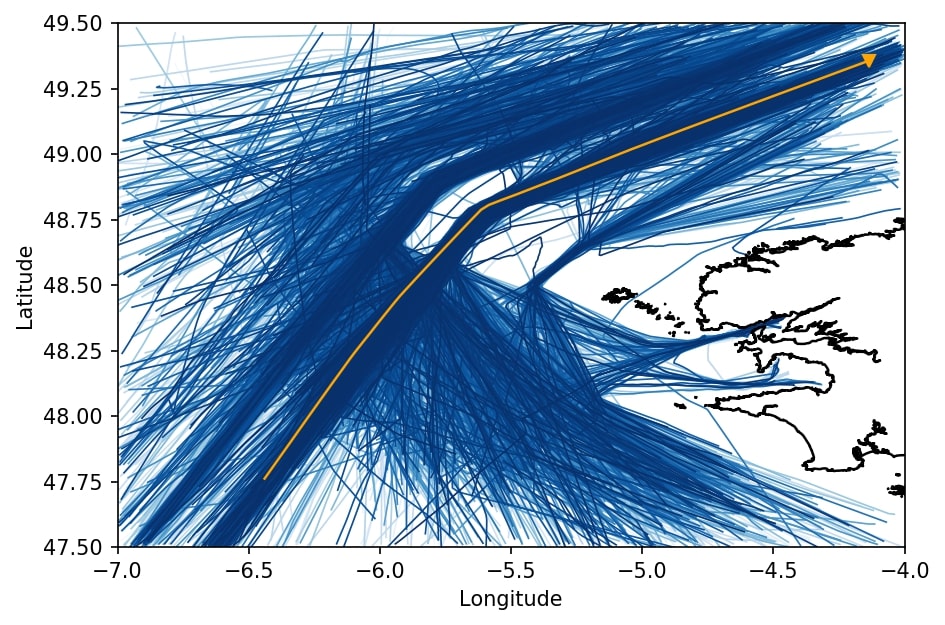}
 	\label{fig:speedcom_a}}%
  \hfil
  	\subfloat[]
  	{\includegraphics[width=\sfsizec]{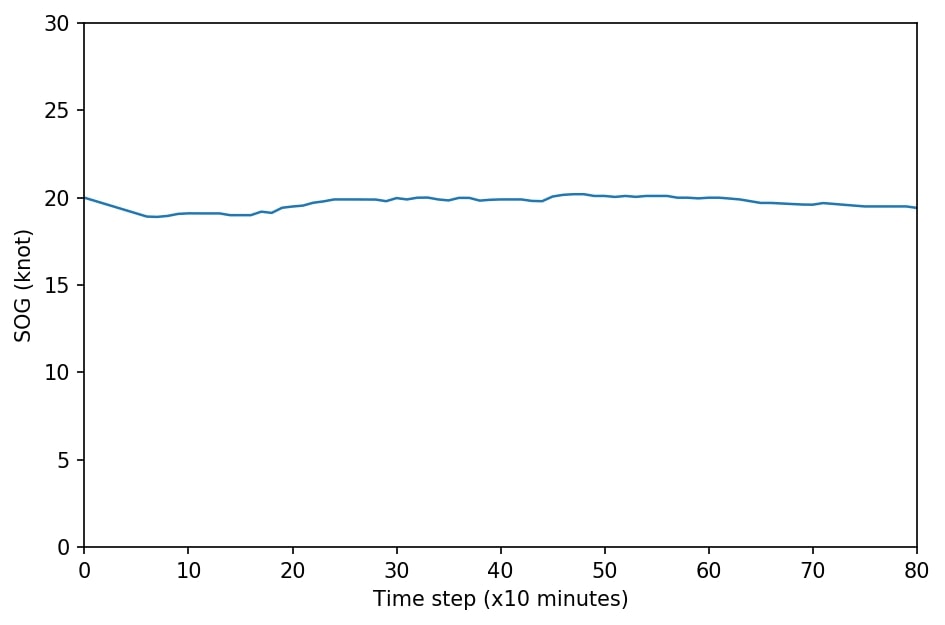}
 	\label{fig:speedcom_b}}%
  \hfil
  	\subfloat[]
    {\includegraphics[width=\sfsizec]{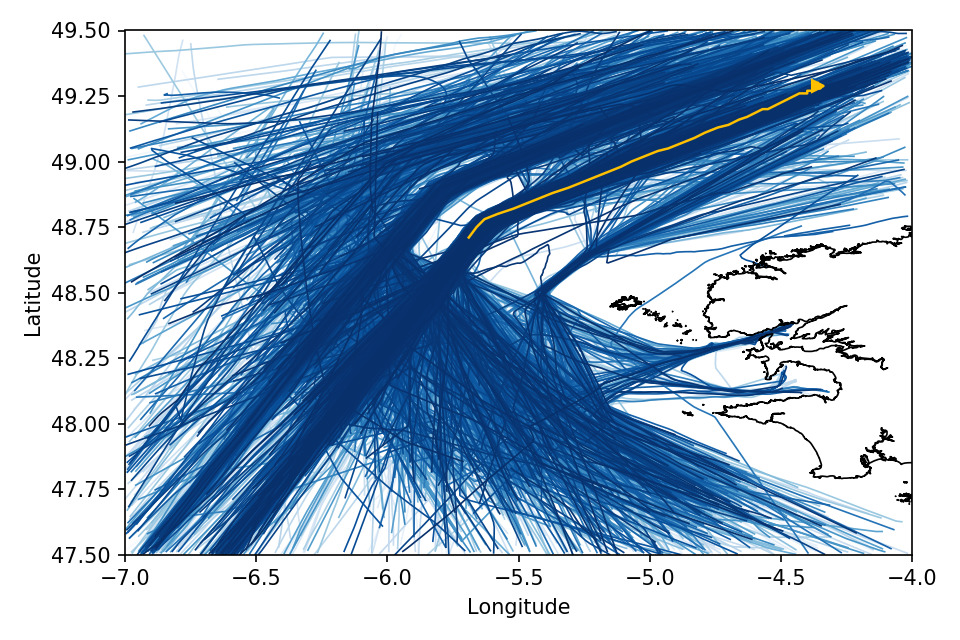}
 	\label{fig:speedcom_c}}%
  \hfil
  	\subfloat[]
    {\includegraphics[width=\sfsizec]{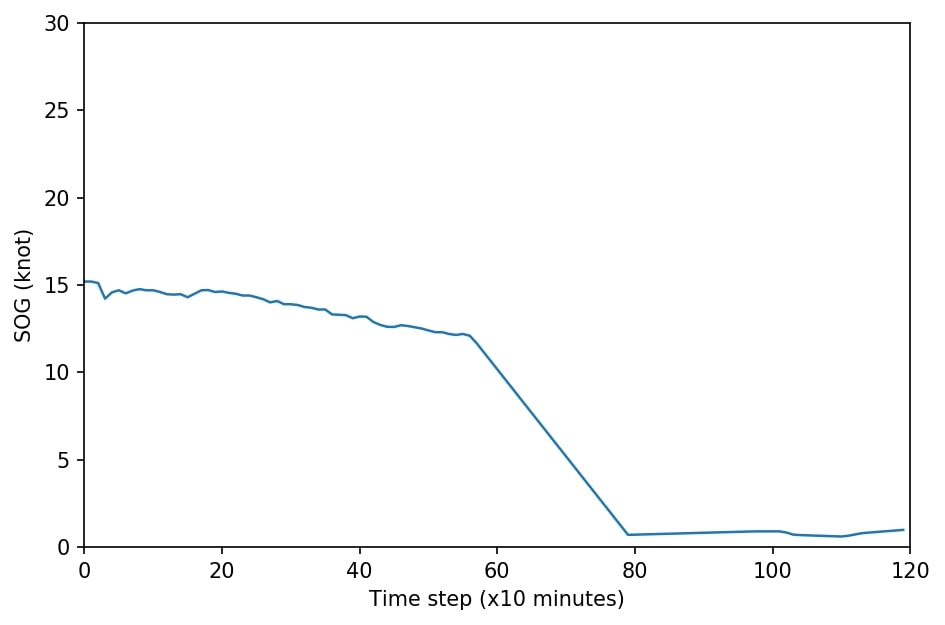}
 	\label{fig:speedcom_d}}%
  \centering
  \caption{Examples of tracks with abnormal speed patterns detected by TREAD and \textit{GeoTrackNet}. (a) An example of a track flagged as abnormal by TREAD and the associated speed pattern (b). The speed of vessels along this route typically varies between 10 and 18 knots while this vessel was moving at around 19 to 20 knots. (c) An example of a track flagged as abnormal by KDE \textit{GeoTrackNet} and the associated speed pattern (d). It involves a sudden slowing-down which may relate to engine problems or abnormal sea/traffic conditions.}
  \label{fig:speedcom}
\end{figure}

\begin{figure}
    \centering
    \includegraphics[width=65mm]{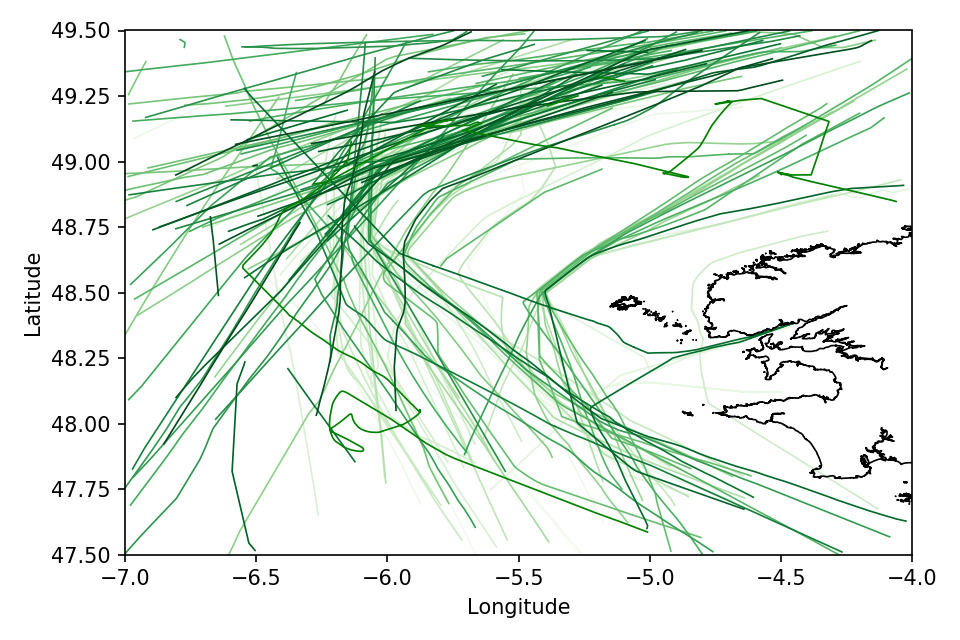}
    \caption{AIS tracks that cannot be mapped to maritime routes, hence cannot be monitored by DBSCAN-based methods. In the test set that comprises only cargo and tanker vessels (from March 21 to March 31, 2017), such tracks account for 13\% of all AIS tracks.}
    \label{fig:DBSCAN_noise}
\end{figure}

The detection of abnormal tracks which do not follow any maritime route like those in Fig. \ref{fig:graph} and Fig. \ref{fig:transhipment} is
a key advantage of \textit{GeoTrackNet} over DBSCAN-based models. Because those tracks can not be mapped to any maritime route, DBSCAN-based methods have two options, either to flag all of them as abnormal or to not monitor them. Since the number of those tracks is high, typically from 10\% to 60\% of the total tracks in the ROI, \cite{pallotta_vessel_2013} (see Fig. \ref{fig:DBSCAN_noise}), neither of these options is relevant for maritime surveillance. 

\begin{figure*}
  \centering
  	\subfloat[]
    {\includegraphics[width=\sfsize]{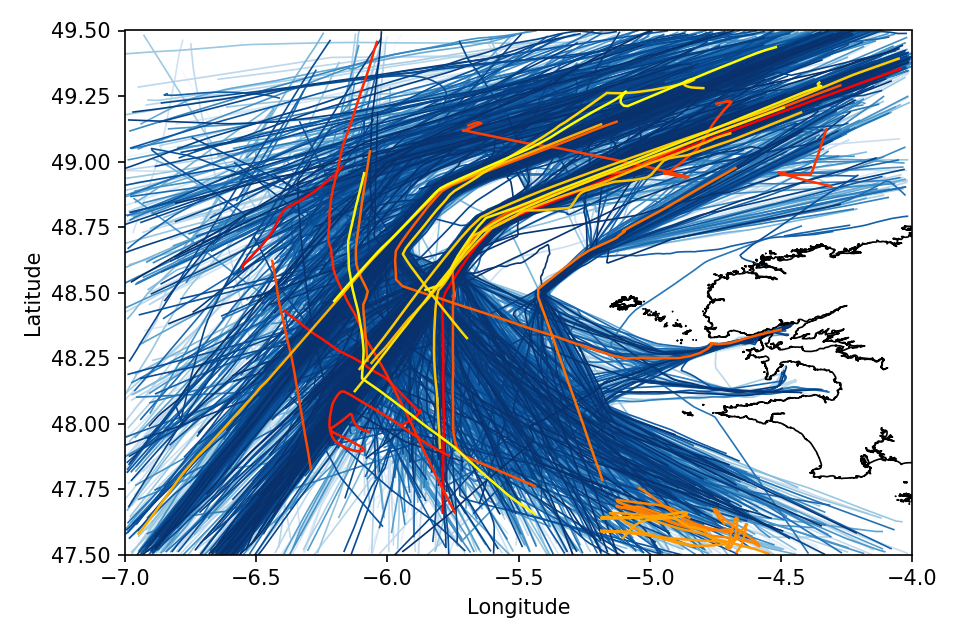}
 	\label{fig:no4hot}}%
    \hfil
  	\subfloat[]
    {\includegraphics[width=\sfsize]{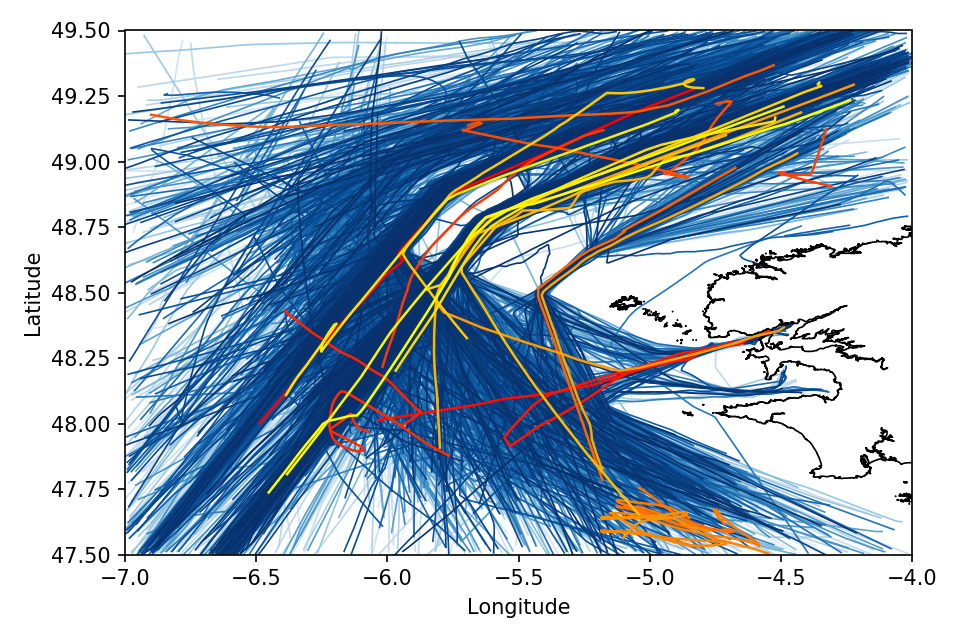}
 	\label{fig:4hot002}}%
    \hfil
  	\subfloat[]
    {\includegraphics[width=\sfsize]{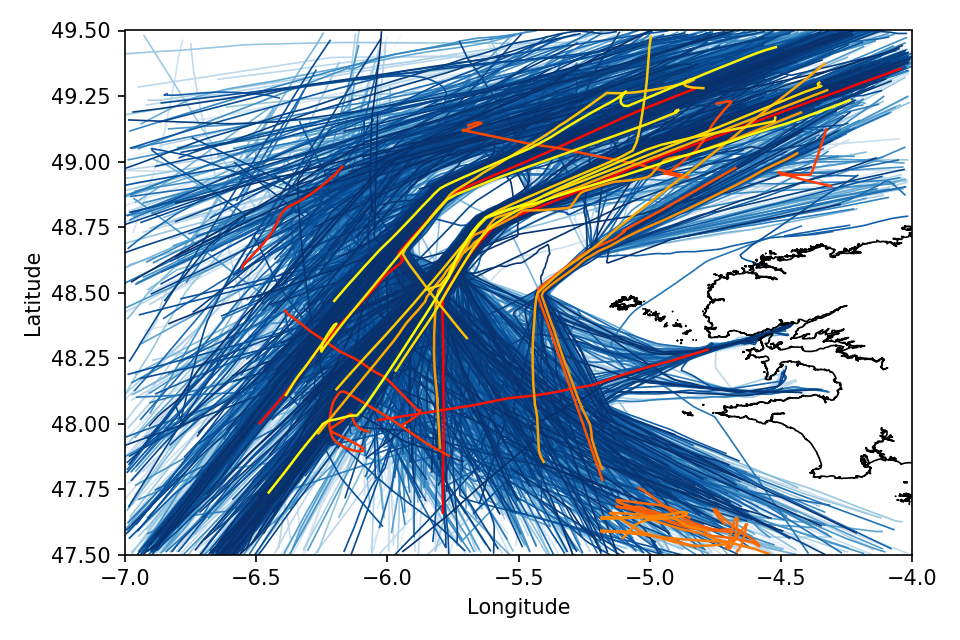}
 	\label{fig:4hot05}}%
 	\hfil
  	\subfloat[]
    {\includegraphics[width=\sfsize]{figures/jan_mar_2017/ct_2017010203_10_20_kde.png}}
    
  \centering
  \caption{\rv{Illustration of the relevance of the ``four-hot" representation. (a) Abnormal tracks detected by a model without the ``four-hot" representation; (b) Abnormal tracks detected by a \textit{GeoTrackNet} model with the resolutions of the latitude, longitude, SOG and COG set to 0.002\degree (=0.2 times the reference setting), 0.002\degree, 1 knot and 5\degree, respectively; (c) Abnormal tracks detected by a \textit{GeoTrackNet} model with the resolutions of the latitude, longitude, SOG and COG set to 0.05\degree (=5 times the reference setting), 0.05\degree, 1 knot and 5\degree, respectively. (d) The reference result, the resolutions of the latitude, longitude, SOG and COG were set to 0.01\degree, 0.01\degree, 1 knot and 5\degree, respectively.}
  \label{fig:ablation_study}}
\end{figure*}

\textbf{Relevance of the ``four-hot" representation}: to demonstrate the relevance of the ``four-hot" representation, we tested the proposed model without the ``four-hot" representation. The result is shown in Fig. \ref{fig:no4hot}. The model fails to detect small, yet very unusual deviations from the common behaviours, such as the double U-turn in Fig. \ref{fig:2Uturn}, or the abnormal turns of the red track in Fig. \ref{fig:turns}. We also tested \textit{GeoTrackNet} with different resolutions of the ``four-hot vector". In general, \textit{GeoTrackNet} is relatively robust to the considered resolutions for the latitude, longitude, SOG and COG. The performance of the model was consistent when we increased or decreased the resolutions of the latitude and the longitude by a factor of 2. When we increased or reduced those resolutions by a factor of 5, the detection started changing. The results with those settings are shown in Fig. \ref{fig:4hot002} and Fig. \ref{fig:4hot05}. When the resolution is too fine, the amount of information that the model has to learn is too much. For example, a spatial resolution of 0.002$\degree$ means that the model has to be able to predict the next position a vessel in 10 minutes (the time resolution of the model) with a tolerance of only 200 meters. On the other hands, if the resolution is too coarse, the information available to the model may not be enough to characterise the movement patterns. For example, a spatial resolution of 0.05$\degree$ means that two positions within a radius of 5 kilometers are not distinguishable.

\textbf{Vessel types}: Another advantage of \textit{GeoTrackNet} is the possibility of applying to any type of vessels. The first step of DBSCAN-based methods is to cluster AIS tracks into maritime routes and learn the signature of each route. Hence, those methods can only apply to vessels that follow maritime routes, i.e. cargo and tanker vessels. By contrast, our method does not impose any hypothesis of this type, so it can apply to any type of vessels. We tested our model on a dataset that comprises all kinds of vessels, the results are shown in Fig. \ref{fig:2017010203_fishing_all}. Since the number of vessels of other types than cargo and tanker is significant, applying the surveillance on all types of vessels is of interest. However, this is a difficult task. Unlike cargo and tanker vessels, some other types, for example fishing vessels, have very complex moving patterns, the model can hardly learn all of them. Even when the model is able to capture all the dynamics of AIS tracks, unexpected results are still inevitable, when the statistical anomalies are actually not suspicious (see Fig. \ref{fig:anomaly_fishing}). There is a trade-off between the monitoring capacity and the performance. When monitoring all types of vessels, it is possible that in a small area, there are some patterns that can be learnt and others that can not. The distribution $p^{C_i}$ is not unimodal anymore. Hence, it cannot be approximated by a Gaussian distribution (see Fig. \ref{fig:gauss_vs_kde}). This explains why the non-parametric density estimation using KDE gives better outcomes in those cases. 

\newcommand{\sfsizeh}{64.5mm}%
\begin{figure}
  \centering
  	\subfloat[]
    {\includegraphics[width=\sfsizeh]{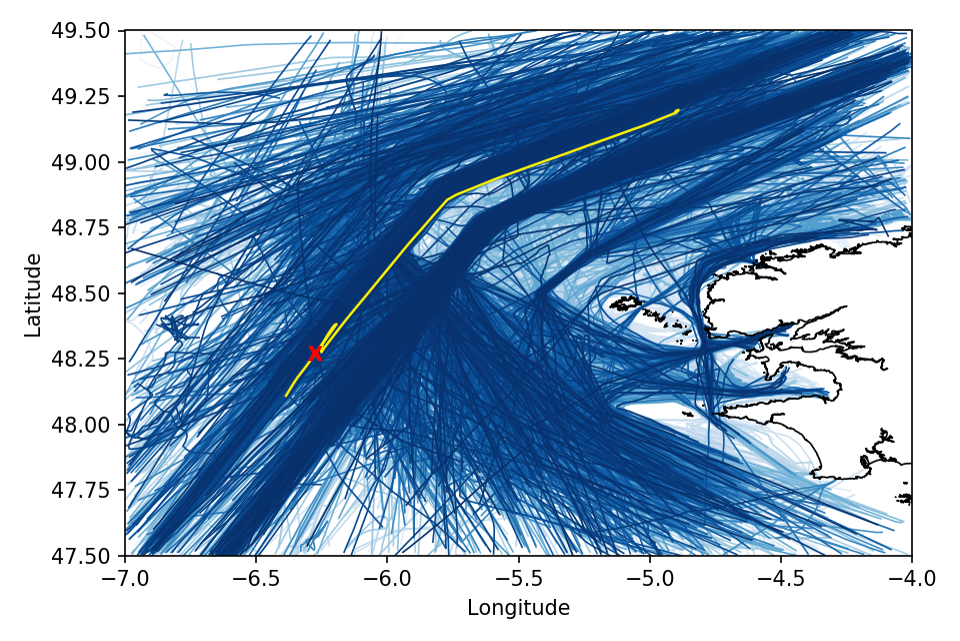}
    \label{fig:gauss_vs_kde_a}}%
  \hfil
  	\subfloat[]
    {\includegraphics[width=\sfsizeh]{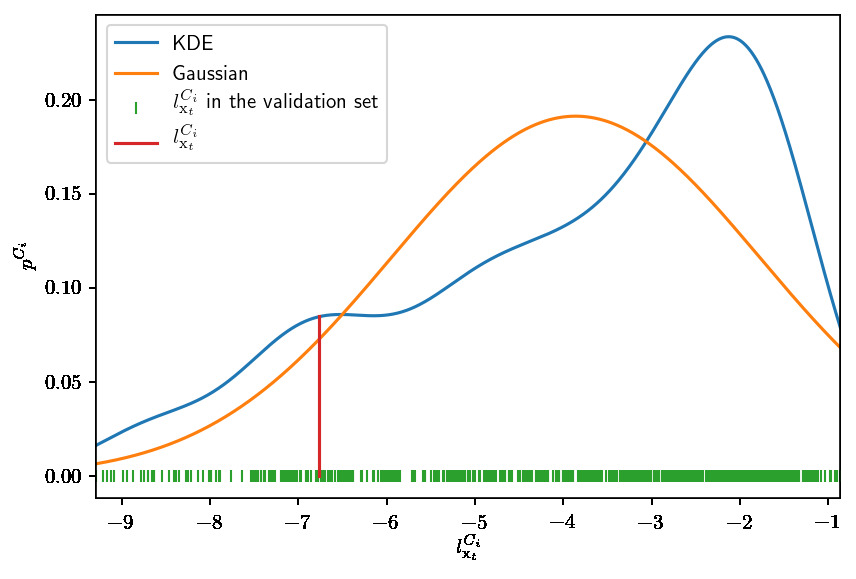}
 	\label{fig:gauss_vs_kde_b}}%
  \centering
  \caption{Comparison between the Gaussian approximation and KDE for distribution $p^{C_i}$.  (a) a track detected as abnormal by KDE \textit{GeoTrackNet}, and not by Gaussian \textit{GeoTrackNet} when the dataset comprises all types of vessels. (b) $p^{C_i}$ of the area around the point ``x" in (a). $p^{C_i}_{KDE}(\mathrm{L} < l_{\vectt{x}}^{C_i}) = 0.128$ while $p^{C_i}_{Gauss}(\mathrm{L} < l_{\vectt{x}}^{C_i}) = 0.082$. Overall, when the data comprises all types of vessels, $p^{C_i}$ is not unimodal and KDE shall be preferred.}
  \label{fig:gauss_vs_kde}
\end{figure}

\newcommand{\sfsizen}{64.5mm}%
\begin{figure}
  \centering
  	\subfloat[]
    {\includegraphics[width=\sfsizen]{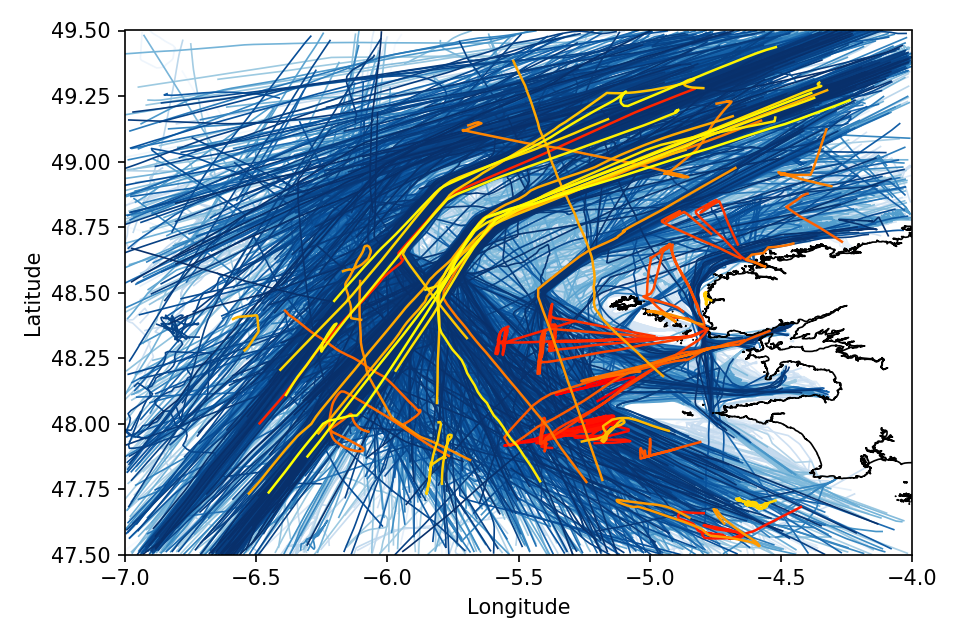}
 	\label{fig:anomaly_fishing}}%
  \hfil
  	\subfloat[]
    {\includegraphics[width=\sfsizen]{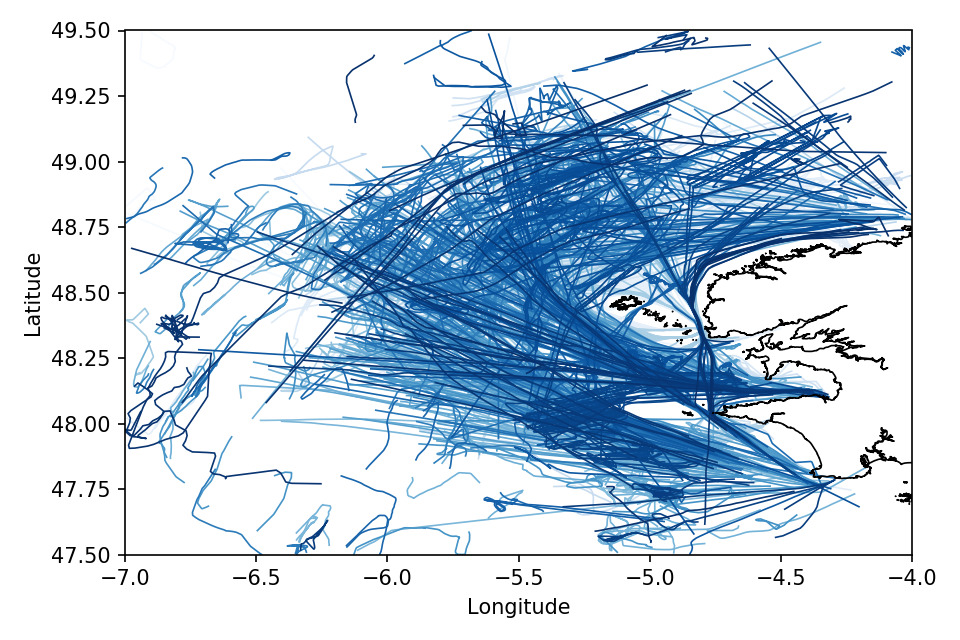}
 	\label{fig:fishing}}%
  \centering
  \caption{Anomaly detection examples of KDE \textit{GeoTrackNet} with AIS data comprising all vessel types from January to March 2017. (a) AIS tracks that are flagged as abnormal by KDE \textit{GeoTrackNet}. Some tracks are statistically abnormal, however, their behaviours are not suspicious. For examples, the red tracks that steamed from land are fishing vessels went fishing; they are detected as abnormal because there are not enough similar AIS tracks in the training set. (b) AIS tracks of fishing vessels in the training set (about 13\% of tracks in the training set).}
  \label{fig:2017010203_fishing_all}
\end{figure}

Hereafter in this paper, unless specified otherwise, the reported results are the results of KDE \textit{GeoTrackNet}.

\textbf{Seasonal effects}: We conducted additional experiments to demonstrate the consistency of \textit{GeoTrackNet}. In this test, the models learnt from the training set of one period were evaluated on the test set of another period\footnote{In real-life applications, we always train the model on recent data. This setting is just to test the consistency of the model}.  Table \ref{tab:ll} shows the average log likelihood on different test sets of models trained on data from January 1 to March 10, 2017. The test sets are data from the 21st to the end of the corresponding month. 
Seasonal effects are small for cargo and tanker vessels. Over seasons, most of the changes are in speed. While for other types of vessels, especially for fishing vessels, the behaviours change completely. That explains why the log likelihood of the model trained on all vessels, from January 1 to March 10, 2017 is considerably low on the test set of September 2017. As shown in Fig. \ref{fig:seasonal_effects}, between winter and summer, the fishing patterns are very different. A model trained on data in one season may not apply to data in another season. These experiments suggest considering season-specific models and/or training a general model which also takes into account a seasonal information.

\begin{table}[!t]
  \renewcommand{\arraystretch}{1.3}
  \caption{Average log likelihood of \textit{GeoTrackNet} for different test sets when trained on AIS data from Jan 1 to Mar 10, 2017.}
  \label{tab:ll}
  \centering
  \begin{tabular}{| l | c | c |}
    \hline
    Test set		& Cargoes and tankers 		& \hphantom{000} All types \hphantom{000} 		\\
    \hline
	March 2017		&  -5.83	& -6.53     	 \\
    September 2017  &  -5.93	& -7.43 	     \\
    March 2018    	&  -5.84    & -6.76 		 \\
    \hline
  \end{tabular}
\end{table}

\newcommand{\sfsizeb}{65mm}%
\begin{figure}
  \centering
  	\subfloat[]
    {\includegraphics[width=\sfsizeb]{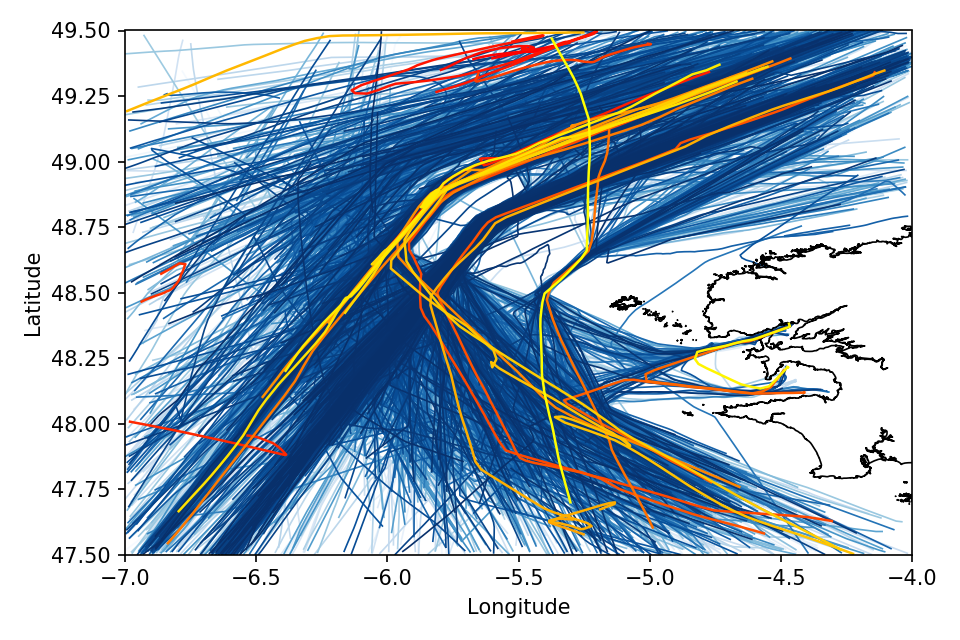}
 	\label{fig:season_ct}}%
  \hfil
  	\subfloat[]
    {\includegraphics[width=\sfsizeb]{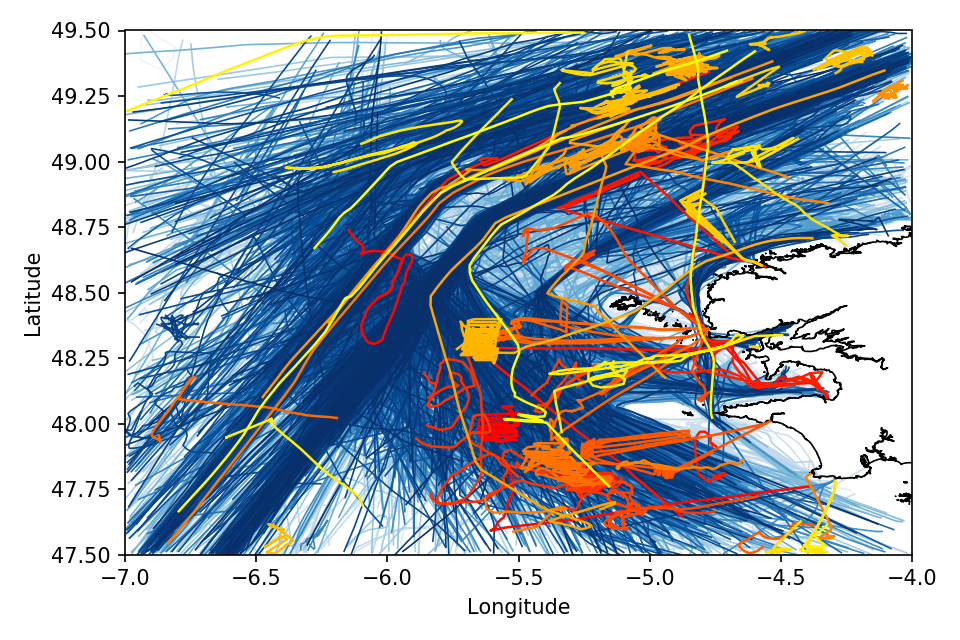}
 	\label{fig:season}}%
  \centering
  \caption{Anomaly detection examples of the model trained on data from January 1 to March 20, 2017 
  and tested on data from July 21 to September 30, 2017.  
  (a) When the data comprise only cargo and tanker vessels. (b) When the data comprise all kind of vessels.}
  \label{fig:seasonal_effects}
\end{figure}

\begin{figure}[]
  \centering
  	\subfloat[]
 	{\includegraphics[width=65mm]{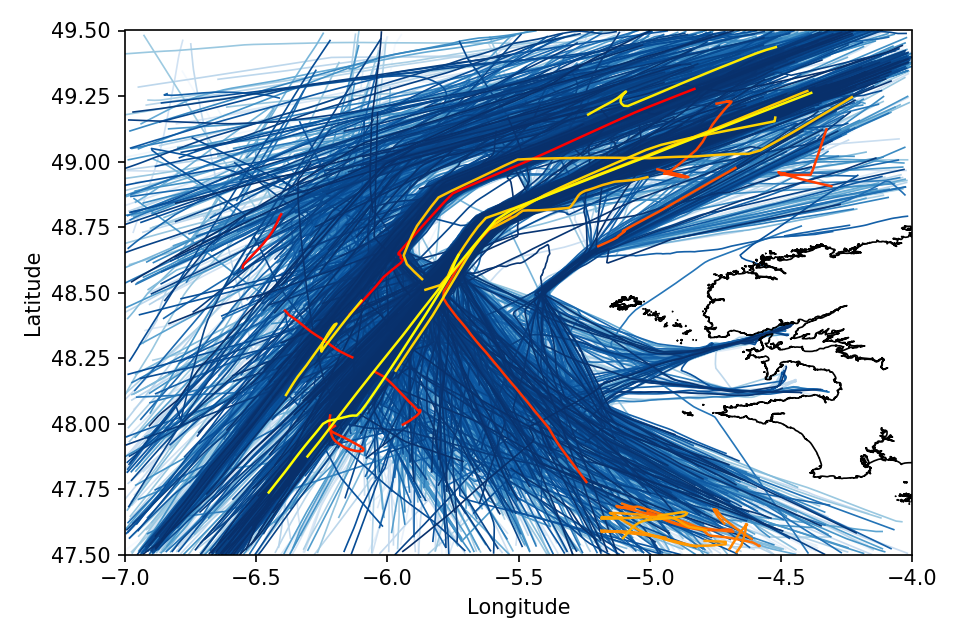}
 	\label{fig:8h}}%
  \hfil
  	\subfloat[]
 	{\includegraphics[width=65mm]{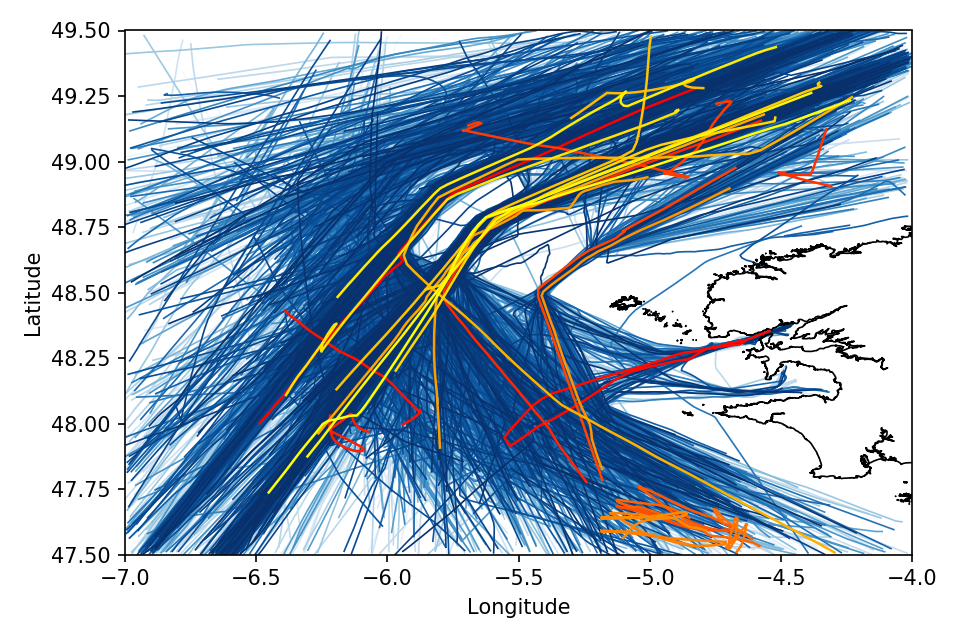}
 	\label{fig:16h}}%
  \centering
  \caption{Effect of the size of the historical data. (a) The maximum duration of each track is 8h; (b) The maximum duration of each track is 16h. If the system does not keep the track long enough, some anomalies may be missed.}
  \label{fig:track_duration}
\end{figure}

\textbf{AIS memory requirements}: In operational mode, one question arises is how long we should keep the past data of each AIS track. In the offline version of \textit{GeoTrackNet}, this quantity is the maximum duration $L_{max}$ of each track. Fig. \ref{fig:track_duration} shows the results of the detection when we split long voyages into small tracks from 4h to: (a) 8h and (b) 16h. Discarding old AIS messages may save memory resources of the system, however, in some cases, we have to observe the track long enough to recognise the anomaly. For example, the voyage of the cargo vessel in Fig. \ref{fig:transhipment} was not detected if the maximum duration of each track is 8h. This is because without knowing the other parts, each segment of this voyage is normal. For dataset presented in this paper, $L_{max} = 16h$ and $L_{max} = 24h$ give the same outcomes. We chose $L_{max} = 24h$ in our experiments as our computational resources could store and process the resulting datasets.

\vspace{-0.7em}
\section{Conclusions and Future work}
\label{sec:conclusions}

We introduced a new approach for maritime anomaly detection using AIS data. To our knowledge, this is the first model which relies on a normalcy model of AIS tracks using a deep learning generative scheme.
The proposed model is novel, both in the way the normalcy model is built and the way deviations from the normalcy are evaluated. More precisely, we exploit Variational Recurrent Neural Networks to represent AIS tracks probabilistically using an original four-hot encoding of  AIS data. Once the approximate distribution of the data is learnt, a geospatial \textit{a contrario} detector is used to evaluate how likely an AIS track is. This detector takes into account the fact that the performance of the learning is geographically dependent. The general idea is that if an AIS message has its log probability lower than other messages' in the same region, it should be flagged as abnormal. An AIS track is abnormal if there are many abnormal messages in this track. 

The key features of the proposed approach are as follows:
\begin{itemize}
    \item It requires a minimal prior knowledge about the data. The model can be applied in different regions without major modifications.
    \item It does not require important hyperparameters such as the number of points in a cluster when using DBSCAN, the number of modes in mixture models, etc.
    \item We can control the percentage of the activities expected to be flagged as abnormal by simply changing the value of $\varepsilon$ in Eq. \eqref{eq:thresholding}.
    \item DBSCAN-based models cannot monitor AIS tracks that do not follow maritime routes. Fig. \ref{fig:DBSCAN_noise} and Fig. \ref{fig:fishing} show that the number of those tracks are significant\footnote{The original paper \cite{pallotta_vessel_2013} reported the fraction of processable AIS messages varied from 40 to 95\%}. Our method  applies to all AIS tracks in the processed area.
    \item The proposed model can detect both geometric/geographic and speed-related anomalies.
    \item The nature of VRNN provides an additional means to condition the output onto external forcing variables or other sources of information. Hence, our model could further benefit from complementary information such as weather conditions,  ocean current situations, etc. Mathematically, it comes to modelling $p(\vectt{x}|\vect{x}_{1:t-1}) = p(\vectt{x}|\vectt{h},\vectt{u})$ with $\vectt{u}$ the forcing variables and additional information.
    \item It is worth noting that anomaly detection is one task (and the most important one) in maritime surveillance. A model that can be integrated into a bigger system would optimise computational and storage resources. In the preliminary version of this  work \cite{nguyen_multi-task_2018}, we showed the proposed NN architecture to be generic and relevant to address other tasks besides anomaly detection such as vessel type recognition and trajectory interpolation. We let the reader to  \cite{nguyen_multi-task_2018} for additional information.
    Regarding computational requirements, the resolution of \textit{GeoTrackNet} is 10 minutes, i.e. the system keeps only one AIS message each 10 minutes. This reduces significantly the amount of data to process and store (by convention, the transmit rate of dynamic AIS message is from every few seconds to every few minutes \cite{imo_international_nodate}). Once the model is learnt, we do not need to store the training dataset. 
    For example, the training set used in this paper from January 1 to March 10, 2017 comprises about 3.3 million AIS messages, which amounts to $\sim$450MB in *.csv format. The learnt model (i.e., VRNN weights) can be embedded into $\sim$40MB in Tensorflow format, which is relatively small. The development of a stream-based version \textit{GeoTrackNet} in \cite{nguyen_detection_2020} supports its relevance for 
    a real-time implementation within a 
    big data and distributed system.
\end{itemize}{}

Although deep learning has recently grown extremely fast and has become the state-of-the-art approach in many domains \cite{lecun_deep_2015}, its achievements in MSA are surprisingly limited. To the best of our knowledge, this work is the first one that applies unsupervised deep learning to maritime anomaly detection. This work opens new avenues to explore new research directions to complement and/or outperform DBSCAN-based approaches. 
As any unsupervised learning-based model, the proposed approach detects events that are statistically unusual. These events may not involve suspicious actions. Ongoing experiments involve analyses by experts to evaluate the consistency of the detections w.r.t. operational requirements. In this respect, the creation of a reference groundtruth dataset would be highly beneficial to advance the state-of-the-art and make benchmarking experiments quantitative. This is however a complicated task that would require a large collaborative effort. \rv{A more thorough study of the relationship between the resolution of the ``four-hot" vector and the corresponding detection results could facilitate the hyper-parameters selection process when applying the model in different zones.}  The proposed neural network representation provides a flexible and powerful means to learn the distribution of AIS tracks, yet uninterpretable. The model is more suitable for a computer-assisted system (where the final decision is still on the human operator) than a fully automatic system. We may emphasise that this representation is also of interest for other tasks, e.g., AIS track interpolation, vessel type identification, as shown in our preliminary work \cite{nguyen_multi-task_2018}. Future work might benefit from such multi-task settings.

\vspace{-0.7em}
\section{Acknowledgements}
\label{secAcknowledgements}

The dataset used in this paper is provided by Collecte Localisation Satellites (CLS) and Erwan Guegueniat. 

We are thankful to Iraklis Varlamis and Mohammad Etemad for for enlightening discussions on the implementation of DBSCAN.

\def\url#1{}
\bibliographystyle{IEEEtran}
\footnotesize
\vspace{-1.5em}
\bibliography{references}

\begin{thebibliography}{10}
\providecommand{\url}[1]{#1}
\csname url@samestyle\endcsname
\providecommand{\newblock}{\relax}
\providecommand{\bibinfo}[2]{#2}
\providecommand{\BIBentrySTDinterwordspacing}{\spaceskip=0pt\relax}
\providecommand{\BIBentryALTinterwordstretchfactor}{4}
\providecommand{\BIBentryALTinterwordspacing}{\spaceskip=\fontdimen2\font plus
\BIBentryALTinterwordstretchfactor\fontdimen3\font minus
  \fontdimen4\font\relax}
\providecommand{\BIBforeignlanguage}[2]{{%
\expandafter\ifx\csname l@#1\endcsname\relax
\typeout{** WARNING: IEEEtran.bst: No hyphenation pattern has been}%
\typeout{** loaded for the language `#1'. Using the pattern for}%
\typeout{** the default language instead.}%
\else
\language=\csname l@#1\endcsname
\fi
#2}}
\providecommand{\BIBdecl}{\relax}
\BIBdecl

\bibitem{nguyen_multi-task_2018}
D.~Nguyen, R.~Vadaine, G.~Hajduch, R.~Garello, and R.~Fablet,
  ``\BIBforeignlanguage{en}{A {Multi}-task {Deep} {Learning} {Architecture} for
  {Maritime} {Surveillance} using {AIS} {Data} {Streams}},'' in
  \emph{\BIBforeignlanguage{en}{2018 {IEEE} {International} {Conference} on
  {Data} {Science} and {Advanced} {Analytics} ({DSAA})}}, Oct. 2018.

\bibitem{wan_four_2016}
Z.~Wan, J.~Chen, A.~E. Makhloufi, D.~Sperling, and Y.~Chen,
  ``\BIBforeignlanguage{en}{Four routes to better maritime governance},''
  \emph{\BIBforeignlanguage{en}{Nature News}}, vol. 540, no. 7631, p.~27, Dec.
  2016.

\bibitem{nanduri_anomaly_2016}
A.~Nanduri and L.~Sherry, ``Anomaly detection in aircraft data using
  {Recurrent} {Neural} {Networks} ({RNN}).''\hskip 1em plus 0.5em minus
  0.4em\relax Ieee, 2016, pp. 5C2--1.

\bibitem{radford_network_2018}
B.~J. Radford, L.~M. Apolonio, A.~J. Trias, and J.~A. Simpson, ``Network
  {Traffic} {Anomaly} {Detection} {Using} {Recurrent} {Neural} {Networks},''
  \emph{arXiv:1803.10769 [cs]}, Mar. 2018, arXiv: 1803.10769.

\bibitem{song_anomalous_2018}
L.~Song, R.~Wang, D.~Xiao, X.~Han, Y.~Cai, and C.~Shi, ``Anomalous trajectory
  detection using recurrent neural network,'' in \emph{International
  {Conference} on {Advanced} {Data} {Mining} and {Applications}}.\hskip 1em
  plus 0.5em minus 0.4em\relax Springer, 2018, pp. 263--277.

\bibitem{ma_detecting_2018}
C.~Ma, Z.~Miao, M.~Li, S.~Song, and M.-H. Yang, ``Detecting {Anomalous}
  {Trajectories} via {Recurrent} {Neural} {Networks},'' in \emph{Asian
  {Conference} on {Computer} {Vision}}.\hskip 1em plus 0.5em minus 0.4em\relax
  Springer, 2018, pp. 370--382.

\bibitem{bouritsas_automated_2019}
G.~Bouritsas, S.~Daveas, A.~Danelakis, and S.~C. Thomopoulos, ``Automated
  {Real}-time {Anomaly} {Detection} in {Human} {Trajectories} using {Sequence}
  to {Sequence} {Networks},'' in \emph{2019 16th {IEEE} {International}
  {Conference} on {Advanced} {Video} and {Signal} {Based} {Surveillance}
  ({AVSS})}.\hskip 1em plus 0.5em minus 0.4em\relax IEEE, 2019, pp. 1--8.

\bibitem{mazzarella_discovering_2014}
F.~Mazzarella, M.~Vespe, D.~Damalas, and G.~Osio, ``Discovering vessel
  activities at sea using {AIS} data: {Mapping} of fishing footprints,'' in
  \emph{17th {International} {Conference} on {Information} {Fusion}
  ({FUSION})}, Jul. 2014, pp. 1--7.

\bibitem{bomberger_associative_2006}
N.~A. Bomberger, B.~J. Rhodes, M.~Seibert, and A.~M. Waxman, ``Associative
  {Learning} of {Vessel} {Motion} {Patterns} for {Maritime} {Situation}
  {Awareness},'' in \emph{2006 9th {International} {Conference} on
  {Information} {Fusion}}, Jul. 2006, pp. 1--8.

\bibitem{pallotta_vessel_2013}
G.~Pallotta, M.~Vespe, and K.~Bryan, ``\BIBforeignlanguage{en}{Vessel {Pattern}
  {Knowledge} {Discovery} from {AIS} {Data}: {A} {Framework} for {Anomaly}
  {Detection} and {Route} {Prediction}},''
  \emph{\BIBforeignlanguage{en}{Entropy}}, vol.~15, no.~6, pp. 2218--2245, Jun.
  2013.

\bibitem{arguedas_maritime_2018}
V.~F. Arguedas, G.~Pallotta, and M.~Vespe, ``Maritime {Traffic} {Networks}:
  {From} {Historical} {Positioning} {Data} to {Unsupervised} {Maritime}
  {Traffic} {Monitoring},'' \emph{IEEE Transactions on Intelligent
  Transportation Systems}, vol.~19, no.~3, pp. 722--732, Mar. 2018.

\bibitem{dobrkovic_maritime_2018}
A.~Dobrkovic, M.-E. Iacob, and J.~van Hillegersberg,
  ``\BIBforeignlanguage{en}{Maritime pattern extraction and route
  reconstruction from incomplete {AIS} data},''
  \emph{\BIBforeignlanguage{en}{International Journal of Data Science and
  Analytics}}, vol.~5, no.~2, pp. 111--136, Mar. 2018.

\bibitem{chung_recurrent_2015}
J.~Chung, K.~Kastner, L.~Dinh, K.~Goel, A.~Courville, and Y.~Bengio, ``A
  {Recurrent} {Latent} {Variable} {Model} for {Sequential} {Data},'' in
  \emph{Advances in neural information processing systems}, Jun. 2015, pp.
  2980--2988.

\bibitem{fraccaro_sequential_2016}
M.~Fraccaro, S.~r.~K. Sø~nderby, U.~Paquet, and O.~Winther, ``Sequential
  {Neural} {Models} with {Stochastic} {Layers},'' in \emph{Advances in {Neural}
  {Information} {Processing} {Systems}}.\hskip 1em plus 0.5em minus 0.4em\relax
  Curran Associates, Inc., 2016, pp. 2199--2207.

\bibitem{maddison_filtering_2017}
C.~J. Maddison, D.~Lawson, G.~Tucker, N.~Heess, M.~Norouzi, A.~Mnih, A.~Doucet,
  and Y.~W. Teh, ``Filtering {Variational} {Objectives},'' in \emph{Advances in
  {Neural} {Information} {Processing} {Systems}}, May 2017, pp. 6576--6586.

\bibitem{rhodes_maritime_2005}
B.~J. Rhodes, N.~A. Bomberger, M.~Seibert, and A.~M. Waxman, ``Maritime
  situation monitoring and awareness using learning mechanisms,'' in
  \emph{{MILCOM} 2005 - 2005 {IEEE} {Military} {Communications} {Conference}},
  Oct. 2005, pp. 646--652 Vol. 1.

\bibitem{laxhammar_anomaly_2008}
R.~Laxhammar, ``Anomaly detection for sea surveillance,'' in \emph{2008 11th
  {International} {Conference} on {Information} {Fusion}}, Jun. 2008, pp. 1--8.

\bibitem{ristic_statistical_2008}
B.~Ristic, B.~L. Scala, M.~Morelande, and N.~Gordon, ``Statistical analysis of
  motion patterns in {AIS} {Data}: {Anomaly} detection and motion prediction,''
  in \emph{2008 11th {International} {Conference} on {Information} {Fusion}},
  Jun. 2008, pp. 1--7.

\bibitem{mascaro_anomaly_2014}
S.~Mascaro, A.~E. Nicholso, and K.~B. Korb, ``Anomaly detection in vessel
  tracks using {Bayesian} networks,'' \emph{International Journal of
  Approximate Reasoning}, vol.~55, no. 1, Part 1, pp. 84--98, Jan. 2014.

\bibitem{dafflisio_maritime_2018}
E.~d'Afflisio, P.~Braca, L.~M. Millefiori, and P.~Willett, ``Maritime {Anomaly}
  {Detection} {Based} on {Mean}-{Reverting} {Stochastic} {Processes} {Applied}
  to a {Real}-{World} {Scenario},'' in \emph{2018 21st {International}
  {Conference} on {Information} {Fusion} ({FUSION})}, Jul. 2018, pp.
  1171--1177.

\bibitem{kawaguchi_anomaly_2018}
Y.~Kawaguchi, ``\BIBforeignlanguage{en}{Anomaly {Detection} {Based} on
  {Feature} {Reconstruction} from {Subsampled} {Audio} {Signals}},'' in
  \emph{\BIBforeignlanguage{en}{2018 26th {European} {Signal} {Processing}
  {Conference} ({EUSIPCO})}}.\hskip 1em plus 0.5em minus 0.4em\relax Rome:
  IEEE, Sep. 2018, pp. 2524--2528.

\bibitem{forti_anomaly_2019}
N.~Forti, L.~M. Millefiori, P.~Braca, and P.~Willett, ``Anomaly {Detection} and
  {Tracking} {Based} on {Mean}–{Reverting} {Processes} with {Unknown}
  {Parameters},'' in \emph{{ICASSP} 2019 - 2019 {IEEE} {International}
  {Conference} on {Acoustics}, {Speech} and {Signal} {Processing} ({ICASSP})},
  May 2019, pp. 8449--8453.

\bibitem{varlamis_network_2019}
I.~Varlamis, K.~Tserpes, M.~Etemad, A.~S. Júnior, and S.~Matwin, ``A {Network}
  {Abstraction} of {Multi}-vessel {Trajectory} {Data} for {Detecting}
  {Anomalies}.'' in \emph{{EDBT}/{ICDT} {Workshops}}, 2019.

\bibitem{tu_exploiting_2017}
E.~Tu, G.~Zhang, L.~Rachmawati, E.~Rajabally, and G.-B. Huang, ``Exploiting
  {AIS} {Data} for {Intelligent} {Maritime} {Navigation}: {A} {Comprehensive}
  {Survey},'' \emph{IEEE Transactions on Intelligent Transportation Systems},
  2017.

\bibitem{riveiro_maritime_2018}
M.~Riveiro, G.~Pallotta, and M.~Vespe, ``Maritime anomaly detection: {A}
  review,'' \emph{Wiley Interdiscip. Rev. Data Min. Knowl. Discov.}, vol.~8,
  2018.

\bibitem{kazemi_open_2013}
S.~Kazemi, S.~Abghari, N.~Lavesson, H.~Johnson, and P.~Ryman, ``Open data for
  anomaly detection in maritime surveillance,'' \emph{Expert Systems with
  Applications}, vol.~40, no.~14, pp. 5719--5729, Oct. 2013.

\bibitem{zhao_maritime_2019}
L.~Zhao and G.~Shi, ``Maritime {Anomaly} {Detection} using {Density}-based
  {Clustering} and {Recurrent} {Neural} {Network},'' \emph{The Journal of
  Navigation}, vol.~72, no.~4, pp. 894--916, 2019.

\bibitem{ester_density-based_1996}
M.~Ester, H.-P. Kriegel, J.~Sander, and X.~Xu, ``A {Density}-based {Algorithm}
  for {Discovering} {Clusters} a {Density}-based {Algorithm} for {Discovering}
  {Clusters} in {Large} {Spatial} {Databases} with {Noise},'' in
  \emph{Proceedings of the {Second} {International} {Conference} on {Knowledge}
  {Discovery} and {Data} {Mining}}, ser. {KDD}'96.\hskip 1em plus 0.5em minus
  0.4em\relax Portland, Oregon: AAAI Press, 1996, pp. 226--231.

\bibitem{coscia_multiple_2018}
P.~Coscia, P.~Braca, L.~M. Millefiori, F.~A.~N. Palmieri, and P.~Willett,
  ``Multiple {Ornstein}-{Uhlenbeck} {Processes} for {Maritime} {Traffic}
  {Graph} {Representation},'' \emph{IEEE Transactions on Aerospace and
  Electronic Systems}, pp. 1--1, 2018.

\bibitem{dafflisio_detecting_2018}
E.~d’Afflisio, P.~Braca, L.~M. Millefiori, and P.~Willett, ``Detecting
  {Anomalous} {Deviations} {From} {Standard} {Maritime} {Routes} {Using} the
  {Ornstein}–{Uhlenbeck} {Process},'' \emph{IEEE Transactions on Signal
  Processing}, vol.~66, no.~24, pp. 6474--6487, Dec. 2018.

\bibitem{lecun_deep_2015}
Y.~LeCun, Y.~Bengio, and G.~Hinton, ``\BIBforeignlanguage{en}{Deep learning},''
  \emph{\BIBforeignlanguage{en}{Nature}}, vol. 521, no. 7553, pp. 436--444, May
  2015.

\bibitem{uney_data_2019}
M.~Üney, L.~M. Millefiori, and P.~Braca, ``Data {Driven} {Vessel} {Trajectory}
  {Forecasting} {Using} {Stochastic} {Generative} {Models},'' in \emph{{ICASSP}
  2019 - 2019 {IEEE} {International} {Conference} on {Acoustics}, {Speech} and
  {Signal} {Processing} ({ICASSP})}, May 2019, pp. 8459--8463.

\bibitem{bengio_representation_2013}
Y.~Bengio, A.~Courville, and P.~Vincent, ``Representation {Learning}: {A}
  {Review} and {New} {Perspectives},'' \emph{IEEE Transactions on Pattern
  Analysis and Machine Intelligence}, vol.~35, no.~8, pp. 1798--1828, Aug.
  2013.

\bibitem{serban_multiresolution_2016}
I.~V. Serban, T.~Klinger, G.~Tesauro, K.~Talamadupula, B.~Zhou, Y.~Bengio, and
  A.~Courville, ``Multiresolution {Recurrent} {Neural} {Networks}: {An}
  {Application} to {Dialogue} {Response} {Generation},'' \emph{arXiv:1606.00776
  [cs, stat]}, Jun. 2016, arXiv: 1606.00776.

\bibitem{gupta_deep_2017}
A.~Gupta, A.~Agarwal, P.~Singh, and P.~Rai, ``A {Deep} {Generative} {Framework}
  for {Paraphrase} {Generation},'' \emph{arXiv:1709.05074 [cs]}, Sep. 2017,
  arXiv: 1709.05074.

\bibitem{zaheer_latent_2017}
M.~Zaheer, A.~Ahmed, and A.~J. Smola, ``\BIBforeignlanguage{en}{Latent {LSTM}
  {Allocation}: {Joint} {Clustering} and {Non}-{Linear} {Dynamic} {Modeling} of
  {Sequence} {Data}},'' in \emph{\BIBforeignlanguage{en}{International
  {Conference} on {Machine} {Learning}}}.\hskip 1em plus 0.5em minus
  0.4em\relax PMLR, Jul. 2017, pp. 3967--3976.

\bibitem{he_probabilistic_2018}
J.~He, A.~Lehrmann, J.~Marino, G.~Mori, and L.~Sigal, ``Probabilistic video
  generation using holistic attribute control,'' in \emph{Proceedings of the
  {European} {Conference} on {Computer} {Vision} ({ECCV})}, 2018, pp. 452--467.

\bibitem{su_variational_2018}
J.~Su, S.~Wu, D.~Xiong, Y.~Lu, X.~Han, and B.~Zhang, ``Variational recurrent
  neural machine translation,'' \emph{arXiv preprint arXiv:1801.05119}, 2018.

\bibitem{ajay_augmenting_2018}
A.~Ajay, J.~Wu, N.~Fazeli, M.~Bauza, L.~P. Kaelbling, J.~B. Tenenbaum, and
  A.~Rodriguez, ``Augmenting physical simulators with stochastic neural
  networks: {Case} study of planar pushing and bouncing,'' in \emph{2018
  {IEEE}/{RSJ} {International} {Conference} on {Intelligent} {Robots} and
  {Systems} ({IROS})}.\hskip 1em plus 0.5em minus 0.4em\relax IEEE, 2018, pp.
  3066--3073.

\bibitem{goodfellow_deep_2016}
I.~Goodfellow, Y.~Bengio, and A.~Courville, \emph{Deep learning}.\hskip 1em
  plus 0.5em minus 0.4em\relax MIT press, 2016.

\bibitem{hochreiter_long_1997}
S.~Hochreiter and J.~Schmidhuber, ``Long short-term memory,'' \emph{Neural
  computation}, vol.~9, no.~8, pp. 1735--1780, 1997.

\bibitem{chung_gated_2015}
J.~Chung, C.~Gulcehre, K.~Cho, and Y.~Bengio, ``\BIBforeignlanguage{en}{Gated
  {Feedback} {Recurrent} {Neural} {Networks}},'' in
  \emph{\BIBforeignlanguage{en}{International {Conference} on {Machine}
  {Learning}}}, 2015, p.~9.

\bibitem{bishop_pattern_2006}
C.~Bishop, \emph{\BIBforeignlanguage{en}{Pattern {Recognition} and {Machine}
  {Learning}}}, ser. Information {Science} and {Statistics}.\hskip 1em plus
  0.5em minus 0.4em\relax New York: Springer-Verlag, 2006.

\bibitem{nguyen_recurrent_2019}
D.~Nguyen, O.~S. Kirsebom, F.~Frazão, R.~Fablet, and S.~Matwin, ``Recurrent
  {Neural} {Networks} with {Stochastic} {Layers} for {Acoustic} {Novelty}
  {Detection},'' in \emph{{ICASSP} 2019 - 2019 {IEEE} {International}
  {Conference} on {Acoustics}, {Speech} and {Signal} {Processing} ({ICASSP})},
  May 2019, pp. 765--769.

\bibitem{desolneux_gestalt_2008}
A.~Desolneux, L.~Moisan, and J.-M. Morel, \emph{\BIBforeignlanguage{en}{From
  {Gestalt} {Theory} to {Image} {Analysis}: {A} {Probabilistic} {Approach}}},
  ser. Interdisciplinary {Applied} {Mathematics}, S.~S. Antman, L.~Sirovich,
  J.~E. Marsden, and S.~Wiggins, Eds.\hskip 1em plus 0.5em minus 0.4em\relax
  New York, NY: Springer New York, 2008, vol.~34.

\bibitem{rosenblatt_remarks_1956}
M.~Rosenblatt, ``Remarks on some nonparametric estimates of a density
  function,'' \emph{The Annals of Mathematical Statistics}, pp. 832--837, 1956.

\bibitem{parzen_estimation_1962}
E.~Parzen, ``On estimation of a probability density function and mode,''
  \emph{The annals of mathematical statistics}, vol.~33, no.~3, pp. 1065--1076,
  1962.

\bibitem{nguyen_detection_2020}
D.~Nguyen, M.~Simonin, G.~Hajduch, R.~Vadaine, C.~Tedeschi, and R.~Fablet,
  ``Detection of {Abnormal} {Vessel} {Behaviors} from {AIS} data using
  {GeoTrackNet}: from the {Laboratory} to the {Ocean},'' in \emph{21st {IEEE}
  {International} {Conference} on {Mobile} {Data} {Management} ({MDM})}, 2020.

\bibitem{kingma_adam:_2015}
D.~P. Kingma and J.~Ba, ``Adam: {A} {Method} for {Stochastic} {Optimization},''
  in \emph{Proceedings of the {International} {Conference} on {Learning}
  {Representations} ({ICLR})}, 2015.

\bibitem{marchi_novel_2015}
E.~Marchi, F.~Vesperini, F.~Eyben, S.~Squartini, and B.~Schuller, ``A novel
  approach for automatic acoustic novelty detection using a denoising
  autoencoder with bidirectional {LSTM} neural networks,'' in \emph{2015 {IEEE}
  {International} {Conference} on {Acoustics}, {Speech} and {Signal}
  {Processing} ({ICASSP})}, Apr. 2015, pp. 1996--2000.

\bibitem{marchi_non-linear_2015}
E.~Marchi, F.~Vesperini, F.~Weninger, F.~Eyben, S.~Squartini, and B.~Schuller,
  ``Non-linear prediction with {LSTM} recurrent neural networks for acoustic
  novelty detection,'' in \emph{2015 {International} {Joint} {Conference} on
  {Neural} {Networks} ({IJCNN})}, Jul. 2015, pp. 1--7.

\bibitem{su_robust_2019}
Y.~Su, Y.~Zhao, C.~Niu, R.~Liu, W.~Sun, and D.~Pei, ``Robust {Anomaly}
  {Detection} for {Multivariate} {Time} {Series} through {Stochastic}
  {Recurrent} {Neural} {Network},'' in \emph{Proceedings of the 25th {ACM}
  {SIGKDD} {International} {Conference} on {Knowledge} {Discovery} \& {Data}
  {Mining}}, ser. {KDD} '19.\hskip 1em plus 0.5em minus 0.4em\relax New York,
  NY, USA: Association for Computing Machinery, Jul. 2019, pp. 2828--2837.

\bibitem{imo_international_nodate}
IMO, ``International {Convention} for the {Safety} of {Life} at {Sea}
  ({SOLAS}), 1974.''

\end{thebibliography}


\end{document}